\documentclass[11pt]{article}

\date{}
\usepackage{times}

\usepackage[utf8]{inputenc} % allow utf-8 input
\usepackage[T1]{fontenc}    % use 8-bit T1 fonts
\usepackage{hyperref}       % hyperlinks
\usepackage{url}            % simple URL typesetting
\usepackage{booktabs}       % professional-quality tables

\usepackage{amsfonts}       % blackboard math symbols
\usepackage{amsmath}
\usepackage{amsthm}
\usepackage{amssymb}
\usepackage{mathtools}

\usepackage{nicefrac}       % compact symbols for 1/2, etc.
\usepackage{microtype}      % microtypography
\usepackage{xspace}
\usepackage{xcolor}
\usepackage{xfrac}

\usepackage{algorithm}
\usepackage[noend]{algpseudocode}

\usepackage[numbers, compress]{natbib}
\usepackage{caption}
\usepackage{enumitem}

\usepackage{graphicx}
\usepackage{float}

\usepackage{xspace}
\usepackage{multirow}
\usepackage{enumitem}
\usepackage{xfrac}
\usepackage{amsmath,amssymb,amsfonts}
\usepackage{amsthm}
\usepackage{booktabs}
\usepackage{algorithm}

\usepackage{multicol,multirow}
\usepackage{url}
\usepackage{xcolor}
\usepackage[noend]{algpseudocode}

\usepackage[framemethod=TikZ]{mdframed}
\usepackage{xcolor}

\algrenewcommand\algorithmicrequire{\textbf{Input:}}
\algrenewcommand\algorithmicensure{\textbf{Output:}}

\newcommand{\method}{{Racing-CVGP}\xspace}

\title{Racing Control Variable Genetic Programming for Symbolic Regression}
\author{Nan Jiang, Yexiang Xue\\
Department of Computer Science, Purdue University \\
\texttt{\{jiang631,yexiang\}@purdue.edu}}

\begin{document}

\maketitle

\begin{abstract}
% What is the main topic
Symbolic regression, as one of the most crucial tasks in AI for science, discovers governing equations from experimental data. 
Popular approaches based on genetic programming, Monte Carlo tree search, or deep reinforcement learning learn symbolic regression from a fixed dataset. They require massive datasets and long training time especially when learning complex equations involving many variables. 
Recently, Control Variable Genetic Programming (CVGP) has been introduced which accelerates the regression process by discovering equations from designed control variable experiments. 
However, the set of experiments is fixed a-priori in CVGP and we observe that sub-optimal selection of experiment schedules delay the discovery process significantly. 
To overcome this limitation, we propose Racing Control Variable Genetic Programming (\method), which carries out multiple experiment schedules simultaneously. A selection scheme similar to that used in selecting good symbolic equations in the genetic programming process is implemented to ensure that promising experiment schedules eventually win over the average ones. The unfavorable schedules are terminated early to save time for the promising ones. 
We evaluate Racing-CVGP on several synthetic and real-world datasets corresponding to true physics laws. 
We demonstrate that Racing-CVGP outperforms CVGP and a series of symbolic regressors which discover equations from fixed datasets. 

\end{abstract}

\section{Introduction}

Automatically discovering scientific laws from experimental data has been a long-standing aspiration of Artificial Intelligence. Its success holds the promise of significantly accelerating scientific discovery. 
A crucial step towards achieving this ambitious goal is symbolic regression, which involves learning explicit expressions from the experimental data.
Recent advancements in this field have shown exciting progress, including works on genetic programming, Monte Carlo tree search,  deep reinforcement learning and their combinations~\cite{doi:10.1126/science.1165893,DBLP:conf/gecco/VirgolinAB19,guimera2020bayesian,DBLP:conf/iclr/PetersenLMSKK21,DBLP:conf/nips/MundhenkLGSFP21,DBLP:conf/iclr/PetersenLMSKK21,DBLP:journals/corr/abs-2203-08808,DBLP:conf/gecco/HeLYLW22,DBLP:conf/iclr/Sun0W023,DBLP:journals/tmlr/TohmeLY23}.

Despite remarkable achievements, the current state-of-the-art approaches are still limited to learning relatively simple expressions, typically involving only a few independent variables.
The real challenge lies in symbolic regression involving multiple independent variables. 
The aforementioned approaches learn symbolic equations from a fixed dataset. 
As a result, these methods require massive datasets and extensive training time to discover complex equations. 

\begin{figure}[!t]
\centering
\includegraphics[width=0.6\linewidth]{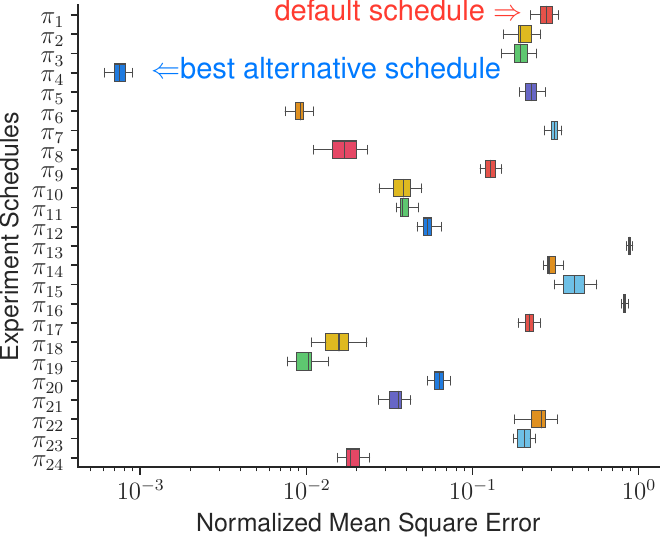}
\caption{Impact of experiment schedules (noted as $\pi$) on learning performance of control variable genetic programming. For the discovery of expression with $4$ variables, there exists a better experiment schedule (\textit{i.e.}, $\pi_4$) among all schedules than the default one (\textit{i.e.}, $\pi_1$), in terms of normalized mean square error.} \label{fig:variable-order-mse}
\end{figure}

Recently, a novel approach called \textit{Control Variable Genetic Programming} (CVGP) has been introduced~\cite{DBLP:arxiv/ecml/ny23} to accelerate symbolic regression. 
Instead of learning from fixed datasets collected a-priori, CVGP carries out symbolic regression using customized control variable experiments. 
As a motivating example, to learn the ideal gas law $pV=nRT$, one can hold $n$ (gas amount) and $T$ (temperature) as constants. It is relatively easy to learn that $p$ (pressure) is inversely proportional to $V$ (volume). 
Indeed, CVGP discovers a chain of simple-to-complex symbolic expressions; e.g., 
first an expression involving only $p$ and $V$, then involving $p$, $V$, $T$, etc. 
In each step, learning is carried out on specially collected datasets where a set of variables are held as constants. 
The major difference between  CVGP and previous approaches is to \textit{actively explore} the space of symbolic expressions via control variable experiments, instead of learning passively from a pre-collected dataset.

However, the set of experiments is fixed a-priori in CVGP. It first learns an equation involving only the first variable, then involving the first two variables, etc.  In particular, CVGP works with a fixed \textit{experiment schedule} (noted as $\pi$), that is the sequences of controlled variables. 
We observe that the sub-optimal selection of experiment schedules delays the discovery process significantly.  See Figure~\ref{fig:variable-order-mse} for a case analysis. 
We run CVGP with all 24 possible experiment schedules and report the quartiles of normalized mean squared errors (NMSE) of the discovered top $20$ expressions.
We see that certain experiment schedules (such as $\pi_4$) are significantly better than others as well as the default experiment schedule $\pi_1$. See more examples in Ablation Study~\ref{sec:ablation}.  Section~\ref{sec:motivation} gives an in-depth explanation.

To overcome this limitation, we propose Racing-CVGP, which automatically discovers good experiment schedules which lead to accurate symbolic regression. 
A selection scheme over the experiment schedules is implemented, similar to that used in selecting good symbolic equations in the genetic programming process, to ensure that promising experiment schedules eventually win over the average ones.
The unfavorable schedules are terminated early to save time for the promising ones.
\method allows for flexible control variables experiments during the discovery process. If a specific set of controlled variable experiments fails to discover a good expression, it is ranked at the bottom and eventually gets removed by the selection scheme. Our idea lets the algorithm avoid spending excessive time on unfavorable experiment schedules and focus on exploring promising controlled variable experiment schedules.

In our experiments, we compare \method against several popular symbolic regression baselines using challenging datasets with multiple variables. On several datasets, we observe that \method discovers expressions with higher quality in terms of NMSE metric against several baselines.  Our \method also takes less computational time than all the baselines.   Since our \method early stops those unfavorable schedules, which commonly leads to a longer time for training.  Notably, our method scales well to expressions with $8$ variables while the GP, CVGP, and GPMeld methods take more than 2 days and thus are time out.

Our contributions can be summarized as follows:
\begin{itemize}%[align=left, leftmargin=0pt, labelwidth=0pt, itemindent=!]
\item We identify a sub-optimal selection of experiment schedule delays the discovery process of symbolic regression greatly. We propose \method to accelerate scientific discovery by maintaining good experiment schedules during learning challenging symbolic regression tasks.
\item Under our racing experiment schedule scheme, a favorable experiment schedule is survived while unfavorable schedules are early stopped. We demonstrate the  time complexity of our \method is approximately close to CVGP, under mild assumptions.
\item We showcase that our \method method leads to faster discovery of symbolic expressions with smaller NMSE metrics, compared to current popular baselines over several challenging datasets.
\end{itemize}

\section{Preliminaries}
\subsection{Symbolic Regression for Scientific  Discovery}
A \textit{symbolic expression} $\phi$ is expressed as variables $\mathbf{x}=\{x_1,\ldots,x_n\}$ and constants $\mathbf{c}=\{c_1,\ldots, c_m\}$, connected by a set of binary operators (like $\{+,-,\times,\div\}$) and/or unary operators (like $\{\sin,\cos, \log,\exp\}$). The operator set is noted as $O_p$.
Each operand of an operator is either a variable, a constant, or a self-contained sub-expression. 
For example, ``$x_1+x_2$'' is a expression with 2 variables ($x_1$ and $x_2$) and one binary operator ($+$).
A symbolic expression can be equivalently represented as a \textit{binary expression tree}, where the leaves nodes correspond to variables and constants and inner nodes correspond to those operators. Figure~\ref{fig:reduced-form}(a) presents an example of the expression tree.

Given a dataset $\mathcal{D}=\{(\mathbf{x}_i, y_i)\}_{i=1}^N$ and a loss function $\ell(\cdot,\cdot)$, the task of \textit{symbolic regression} is to find the optimal symbolic expression $\phi^*$ with minimum loss over the dataset $\mathcal{D}$, among the set of all candidate expressions (noted as $\Pi$):
\begin{equation} \label{eq:objective}
\phi^*\leftarrow\arg\min_{\phi\in \Pi}\;\frac{1}{N} \sum_{i=1}^{N} \ell(\phi(\mathbf{x}_i,\mathbf{c}), y_i),
\end{equation}
where the values of open constants $\mathbf{c}$ in $\phi$ are determined by fitting the expression to the dataset $\mathcal{D}$. The loss function $\ell(\cdot,\cdot)$ measures the distance between the output from the candidate expression $\phi(\mathbf{x}_i,\mathbf{c})\in\mathbb{R}$ and the ground truth $y_i\in\mathbb{R}$. A common choice of the loss function is Normalized Mean Squared Error (NMSE). 
Symbolic regression is shown to be NP-hard~\cite{journal/tmlr/virgolin2022}, due to the exponentially large size of all the candidate expressions $\Pi$. 

\begin{figure*}[!t]
  \centering
    \includegraphics[width=1\linewidth]{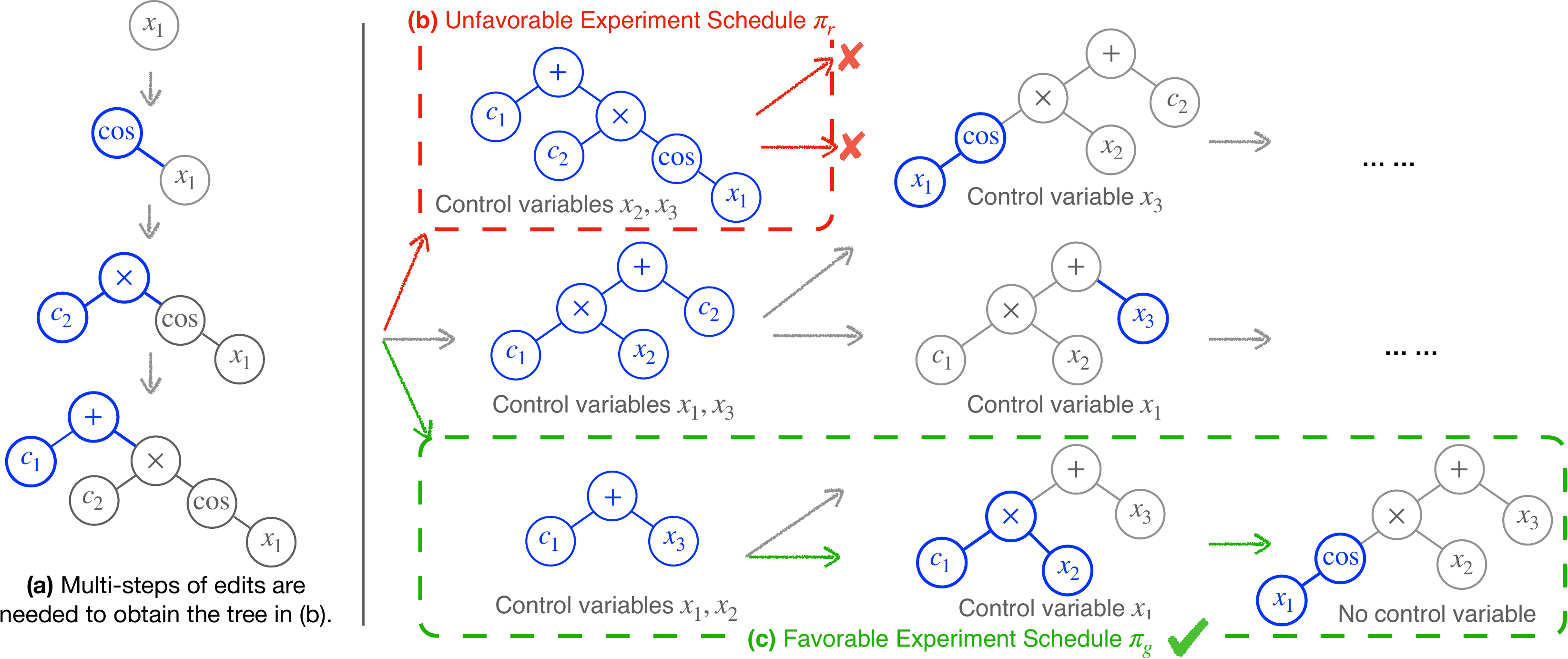}
    \caption{Favorable experiment schedule $\pi_g$ is survived while unfavorable schedule $\pi_r$ is early stopped under our racing experiment schedule scheme.  
    \textbf{(a)} Multiple steps of edits are needed to transform from a randomly initialized expression ``$x_1$'' to a complex expression ``$c_1+c_2\cos(x_1)$''. The newly inserted parts (by genetic programming algorithm) are highlighted in blue. \textbf{(b)} The red experiment schedule $\pi_r$ is unfavorable, since it requires many edits to reach the expression tree in the red box (shown in (a)). The red experiment schedule is thus early stopped. \textbf{(c)} The green experiment schedule $\pi_g$ is promising since it is relatively easy to discover and every change over the expression tree is reasonable. Section~\ref{sec:motivation} gives a detailed explanation.}
    \label{fig:motivation} 
    % \vspace{-1.5em}
\end{figure*}

\noindent\textbf{Genetic Programming for Symbolic Regression.}
Genetic Programming (GP) has emerged as a popular method for solving symbolic regression.
The core idea of GP involves managing a pool of candidate symbolic expressions, noted as $\mathcal{P}$. In each generation, these candidates undergo \textit{mutation} and \textit{mating} steps with certain probabilities. The mutation operations randomly replace, insert a node in the expression tree or delete a sub-tree. 
The mating operations pick a pair of parent expression trees and exchange their two random sub-trees.
Subsequently, during the \textit{selection} step, expressions with the highest fitness scores, are chosen as candidates for the next generation. 
Here the fitness scores (noted as $\mathbf{o}\in\mathbb{R}^N$) indicate the closeness of predicted outputs to the ground-truth outputs, like the negative NMSE.
Over several generations, the expressions that fit the data well, exhibiting high fitness scores, survive in the pool of candidate solutions. The best expressions discovered over all generations are recorded as \textit{hall-of-fame} solutions, noted as $\mathcal{H}$.

\subsection{Control Variable Trials} 
In a regression problem, control variable trials study the relationship between a few input variables and the output with the rest input variables fixed to be the same~\cite{lehman2004designing,hunermund2020nuisance}. This idea was historically proposed to discover natural physical law, known as the BACON system~\cite{DBLP:conf/ijcai/Langley77,DBLP:conf/ijcai/Langley79,DBLP:conf/ijcai/LangleyBS81}.
Recently, this idea is explored in solving multi-variable symbolic regression problems~\cite{DBLP:arxiv/ecml/ny23}, \textit{i.e.}, CVGP. 

\begin{figure}[!t]
  \centering
    \includegraphics[width=1\linewidth]{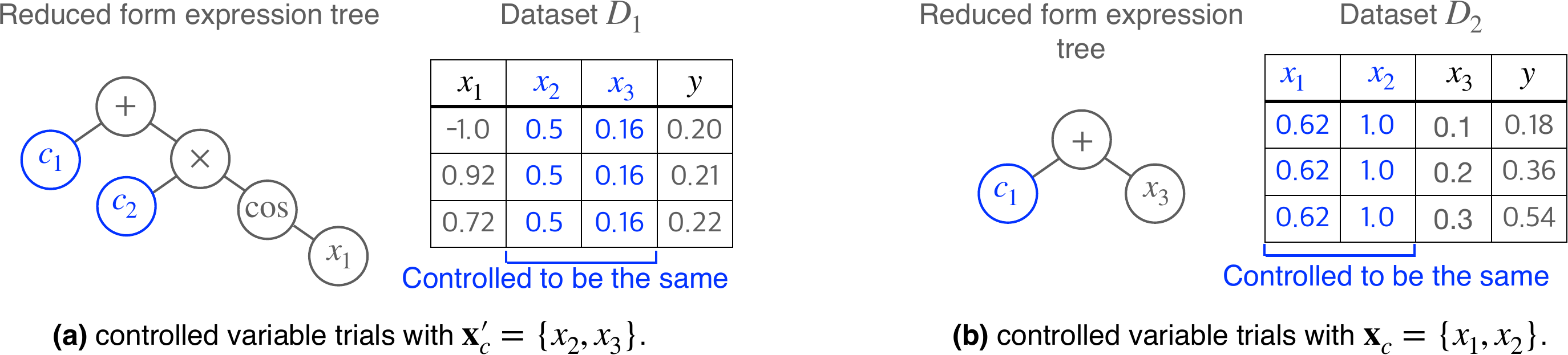}
    \caption{
 \textbf{(a)} When controlling variables $x_2$ and $x_3$, the  ground-truth expression $\phi=x_2\cos(x_1)+x_3$ reduces to $c_1\cos(x_1)+c_2$.
 To fit the open constants $\{c_1,c_2\}$, the dataset sampled from the \texttt{DataOracle} needs to set the value of controlled variables $\{x_2,x_3\}$ to be the same.
\textbf{(b)} Controlling variables $x_1$ and $x_2$ reduces the ground-truth to $c_1x_3$.}
    \label{fig:reduced-form} 
    % \vspace{-1.5em}
\end{figure}

Let $\mathbf{x}_c\subseteq\mathbf{x}$ denote those control variables, and the rest are free variables. Since the values of controlled variables are fixed in each trial, which behave exactly the same as constants for the learning method. In the controlled setting,  the ground-truth expression behaves the same after setting those controlled variables as constants, which is noted as the \textit{reduced form expression}. See Figure~\ref{fig:reduced-form}(a,b) for two {reduced form} expressions with different control variable settings.

For a single control variable trial in symbolic regression,  the corresponding dataset  $\mathcal{D}=\{(\mathbf{x}_i, y_i)\}_{i=1}^m$ is first generated, where the controlled variables are fixed to one value and the rest variables are randomly assigned. That is $\mathbf{x}_{i,k}=\mathbf{x}_{j,k}$ for the control variable $x_k$ ($x_k\in \mathbf{x}_c$) and $1\le i,j\le N$.  See Figure~\ref{fig:reduced-form}(a,b) for example datasets generated from the control variable trials. 
%
% CVGP method is then evoked to search for those optimal {reduced-form} expressions. 
Given a reduced form expression and corresponding dataset, the values of open constants in the expression are determined by gradient-based optimizers, like the BFGS algorithm.  In Figure~\ref{fig:reduced-form}(a), the optimal values of open constants are $c_1=0.5,c_2=0.16$. Similarly in  Figure~\ref{fig:reduced-form}(b), we have $c_1=1.8$. The loss values (defined in Equation~\eqref{eq:objective}) of these two controlled variable trails over the dataset  $D_1$ and dataset $D_2$ are equal to $0$, indicating the optimal fitness scores.

The CVGP is built on top of the above control variable trials and genetic programming algorithms~\cite{DBLP:arxiv/ecml/ny23}. To fit an expression of $n$ variables, CVGP initially only allows only variable $x_1$ to vary and controls the values of all $n-1$ variables (\textit{i.e.}, $\mathbf{x}_c=\mathbf{x}\backslash\{x_1\}$).
Using GP as a subroutine, CVGP finds a pool of expressions $\{\phi_{1}, \ldots, \phi_{N_p}\}$ which best fit the data from this controlled experiment. 
Notice 
$\{\phi_{1}, \ldots, \phi_{N_p}\}$ are restricted 
to contain the only one free variable $x_1$ and $N_p$ is the pool size. This fact 
renders fitting them a lot easier than directly fitting the expressions involving all $n$ variables. A small error implies $\phi_{i}$ is close to the ground truth reduced to the one free variable.  In the 2nd round, CVGP adds a second free variable $x_2$ and starts fitting $\{\phi'_{1}, \ldots, \phi'_{N_p}\}$ using the data from control variable experiments involving the 
 two free variables $x_1,x_2$  (\textit{i.e.}, $\mathbf{x}_c=\mathbf{x}\backslash\{x_1,x_2\}$). After $n$ rounds,  the expressions
in the CVGP pool consider expressions involving all $n$ variables.  Note that CVGP assumes the existence of a \texttt{DataOracle} that allows for query a batch data with specified control variables.

% where $\pi_t$ denotes the set of controlled variables at round $t$. %For the first round, the data generated from the \texttt{DataOracle} is listed as dataset $\mathcal{D}_2$ in Figure~\ref{fig:reduced-form}(b).

\section{Methodology}
We first brief the issue with a fixed experiment schedule for the existing CVGP method in discovering symbolic regression. Then we present our racing experiment schedule for control variable genetic programming, \textit{i.e.}, \method.

\subsection{Motivation} \label{sec:motivation}
We define an \textit{experiment schedule}, noted as $\pi$, as a sequence of variables controlled over all the rounds in CVGP. %$\pi_t$ is the set of control variable at $t$-th round of CVGP.
We use Figure~\ref{fig:motivation} to demonstrate different experiment schedules for the discovery of the ground-truth expression $\phi=\cos(x_1)x_2+x_3$. In Figure~\ref{fig:motivation}(c),  CVGP runs an experiment schedule with control variables $\{x_1,x_2\}$ in the first round and runs with control variables $\{x_1\}$ in the second round and with no variable control $\emptyset$ at the last round. The corresponding experiment schedule is $\pi_g=(\{x_1,x_2\}, \{x_1\}, \emptyset)$. Similarly, Figure~\ref{fig:motivation}(b) shows the default experiment schedule of CVGP that control variables $\{x_2,x_3\}$ initially and then control variable $\{x_3\}$, finally control no variable $\emptyset$, which is noted as $\pi_r=(\{x_2,x_3\}, \{x_3\}, \emptyset)$.

% we change the ordering.
Our key observations are that: (1) The experiment schedule plays a vital impact on the performance of CVGP than other components in the algorithm. (2) Some expressions are much easier to detect for specific experiment schedules. The existing CVGP method only considers a fixed experiment schedule $\pi=(\{x_2,\ldots,x_n\}, \{x_3,\ldots,x_n\},\ldots, \{x_n\}, \emptyset)$ for discovering expression involving $n$ variables. 
This fixed experiment schedule leads to sub-optimal performance of CVGP over some expressions, requiring more training data and computational time than other alternative schedules. 
See Figure~\ref{fig:variable-order-mse} for an empirical evaluation of different experiment schedules over the final identified expressions by the same CVGP method. See more examples in Ablation Study~\ref{sec:ablation}.

In Figure~\ref{fig:motivation}, we use the discovery of an expression $\phi=\cos(x_1)x_2+x_3$ from the Feynman dataset as an example. The alternative (green) experiment schedule $\pi_g$ in Figure~\ref{fig:motivation}(c) is favorable while the default (red) schedule $\pi_r$ in Figure~\ref{fig:motivation}(b) is not.  
In Figure~\ref{fig:motivation}(a), we visualize 3 necessary steps to reach from a randomly initialized expression tree ``$x_1$'' to the final tree ``$c_1+c_2\cos(x_1)$'' in Figure~\ref{fig:motivation}(b). The edited subtrees are highlighted in blue color.
Every step of editing is conducted by the mutations, mating, and selection steps in GP. Every intermediate expression requires drawing batches of training data to fit open constants. The more number of necessary edits over the expression tree, the more time and data used for the 1st round.
In comparison, it takes 1 step of edits in the tree to reach the first expression ``$c_1+x_3$'' in the green experiment schedule, which leads to faster discovery using less training data.  
Following the green experiment schedule $\pi_g$, it takes 1 step of edits to reach the expression at the second round ``$c_1x_2+x_3$'' and the last round ``$\cos(x_1)x_2+x_3$''.
Since every change over the expression tree is reasonable for the alternative (green) experiment schedule $\pi_g$, it is easier for the same GP algorithm to discover the ground-truth expression using less data and time using alternative schedule $\pi_g$ than the default schedule $\pi_g$.

Note that directly evoking CVGP as a subroutine with multiple experiment schedules will not solve the problem. The expression in Figure~\ref{fig:motivation} has $24$ different experiment schedules. The total running time is summarized in Figure~\ref{fig:quartile-time-exp-schedule}. In general, for an expression involving $n$ variables, there are $n!$  exponential many experiment schedules.  It is time-intractable to run  CVGP with all the experiment schedules for real-world scale problems. %

To tackle the above issue, we propose a racing scheme over the experiment schedules. Our main principles are (1) maintaining multiple experiment schedules rather than one, and (2) allowing promising experiment schedules to survive while letting unfavorable schedules early stop. 
Our \method has a much higher chance to detect high-quality expression using less training data and computational time than the existing CVGP. 

Specifically, we implement a schedule selection procedure. Every expression in the population pool $\phi\in\mathcal{P}$ is attached with its own experiment schedule.
In each round, we execute GP over all the expressions in the population pool for several generations.
At the end of every round, the racing selection scheme removes (\textit{resp.} preserves) those expressions with bad (\textit{resp.} good) experiment schedules, based on their fitness scores. 
So that those schedules that lead to higher fitness scores have a higher probability of survival.

We use Figure~\ref{fig:motivation} to visualize the process of our \method. We first initialize the population pool $\mathcal{P}$  in GP with several expressions for each control variable setting. We randomly generate simple expressions involving only $x_1$ with the control variables being $\{x_2,x_3\}$, where every expression is attached with a (partial) experiment schedule $\pi=(\{x_2,x_3\})$. We repeat this random expression generation for all the rest $n-1$ control variable settings.
% Given control variable  $\mathbf{x}_c=\mathbf{x}\setminus\{x_i\}$, we generate simple expressions only involve $x_i$, for $1\le i\le 3$.
For the 1st round, the GP algorithm is evoked over the population pool for several generations.  Then we rank the expressions in the pool by the fitness score of the expression, where those expressions with higher fitness scores rank at the top of the pool. We only preserve top $N_p$ expressions in population pool $\mathcal{P}$. Since it is much easier to detect $c_1+x_3$ under control variable $\{x_1,x_2\}$ setting, the preserved majority expressions are attached with the experiment schedule $\pi_1=\{x_1,x_2\}$. 
This ensures that we early stop the unfavorable experiment schedule $\pi=\{x_2,x_3\}$ in Figure~\ref{fig:motivation}(b). Prior to the 2nd round, we randomly set free one variable from $\pi_1$. Figure~\ref{fig:motivation}(c) set the free variable $x_2$ and only variable $x_1$ is controlled in the 2nd round. In the 3rd round, the majority of the expressions in the population is attached to the experiment schedule $\pi_g=(\{x_1,x_2\},\{x_1\},\emptyset)$, since every change over the expression tree is reasonable. 
The total computational resources are saved from spending time searching for the expression tree in Figure~\ref{fig:motivation}(b) to explore expressions with experiment schedule  $\pi=(\{x_1,x_2\},\{x_1\})$ in Figure~\ref{fig:motivation}(c).

\subsection{Racing Control Variable Genetic Programming}
The high-level idea of  \method is building more complex symbolic expressions involving more and more variables following those promising experiment schedules.

\noindent\textbf{Notations.} $K$ multiple control variable trials can be noted as a tuple  $\langle\phi,\mathbf{o},\mathbf{c},\mathbf{x}_c,\pi, K,\{\mathcal{D}_k\}_{k=1}^K\rangle$.
Here $\phi$ stands for the symbolic expression;  the fitness scores  $\mathbf{o}\in\mathbb{R}^K$ for expression $\phi$ indicates the
closeness of predicted outputs to the ground-truth outputs;  $\mathbf{c}\in\mathbb{R}^{K\times L}$ are the best-fitted values (by gradient-based optimizers) to open constants.
Here $L$ stands for the number of open constants in the expression $\phi$; $\mathbf{x}_c\subseteq \mathbf{x}$  is the set of control variables; $\pi$ is the (partial) experiment schedule that leads to the current expression $\phi$. 
$\mathcal{D}_k=\{(\mathbf{x}_i,y_i)\}_{i=1}^m$ ($1\le k\le K$) is a randomly sampled batch of data  from $\mathtt{DataOracle}$ with control variables $\mathbf{x}_c$. $m$ denotes the batch size.

\noindent\textbf{Initialization.} For single variable $x_i\in\mathbf{x}$, we create a set of candidate expressions that only contain variable $x_i$ and save them into the population pool $\mathcal{P}$.  Then we apply a GP-based algorithm to find the best-fitted expressions, which is referred to as the $\mathtt{BuildGPPool}$ function. The initialization step corresponds to Lines 3-7 in Algorithm~\ref{alg:racing-cvgp}.

\noindent\textbf{Execution Pipeline.} Given the current control variables $\mathbf{x}_c$, we first evoke the $\mathtt{DataOracle}$ to generate data batches  $\mathcal{D}_1,\ldots, \mathcal{D}_K$. %
This corresponds to changing experimental conditions in real science experiments. 
We then fit open constants in the candidate expression $\phi_{new}$ with the data batches by gradient-based optimizers like BFGS~\cite{fletcher2000practical}. This step is noted as the $\mathtt{Optimize}$ function. Then we obtain the fitness score vector $\mathbf{o}$ and solutions to open constants $\mathbf{c}$. 
We save the tuple $\langle\phi_{new},\mathbf{o}, \mathbf{c}, \pi, \mathbf{x}_c\rangle$ into new population pool $\mathcal{P}_{new}$. This step corresponds to  Lines 9-12 in Algorithm~\ref{alg:racing-cvgp}. 

Then GP algorithm is applied for $\#\mathtt{Gen}$ generations to search for optimal structures of the expression trees in the population pool $P_{new}$. The function $\mathtt{GP}$ is a minimally modified genetic programming algorithm for symbolic regression, which is detailed in Appendix~\ref{apx:gp}. We only preserve $N_p$ best expressions in the population $\mathcal{P}$. Note that every expression is evaluated with the different data from \textit{its own control variables}. An unfavorable (partial) experiment schedule will be removed at this step when the corresponding expression $\phi$ has a low fitness score. The schedules in the pruned population pool $\mathcal{P}$ indicate they are favorable. 

Key information is obtained by examining the outcomes of $K$-trials control variable experiments:
(1) consistent close-to-zero fitness value, implies the fitted expression is close to the ground-truth equation in the reduced form. 
That is $\sum_{k=1}^K\mathbb{I}(o_k\le \varepsilon)$ should equal to $K$, where $\mathbb{I}(\cdot)$ is an indicator function and  $\varepsilon$ is the threshold for the fitness scores.
(2) Given the equation is close to the ground truth, an open constant having similar best-fitted values across $K$ trials suggests the open constants are stand-alone.  Otherwise, that open constant is a \textit{summary} constant, that corresponds to a sub-expression involving those control variables $\mathbf{x}_c$.
The $j$-th open constant is an standalone constant when $\mathbb{I}(\mathtt{var}(\mathbf{c}_{j})\le \varepsilon')$ is evaluated to $1$, where $\mathtt{var}(\mathbf{c}_{j})$ indicates the variance of the solutions for $j$-th open constant. It is computed as $\sum_{k=1}^K(c_{k,j}-\frac{1}{K}\sum_{k=1}^Kc_{k,j})^2$. 
Hyper-parameter $ \varepsilon'$ is the threshold. The above steps are noted as $\mathtt{FreezeEquation} $ function (in Line 16 of Algorithm~\ref{alg:racing-cvgp}). This freeze operation reduces the search space and accelerates the discovery process. Examples are available in Figure~\ref{fig:freeze} in Appendix.

 % $\mathbf{c}_j\in\mathbb{R}^K$

Finally, we randomly drop a control variable in $\mathbf{x}_c$ and update the schedule $\pi$ for each expression $\phi$ in the population pool $\mathcal{P}$. At $i$-th round, the size of all control variable for all the expressions equals to $n-i$. After $n$ rounds, we return the expressions in hall-of-fame $\mathcal{H}$ with the best fitness values over all the schedules. Expressions in $\mathcal{H}$ are evaluated using the same data with no variable controlled.

\begin{algorithm}[!t]
   \caption{Racing Control Variable Genetic Programming}\label{alg:racing-cvgp}  
   \begin{algorithmic}[1]
   \Require{\#experiment trials $K$;  operator set $O_p$; ground-truth expression $\phi_{gt}$; \#of input variables $n$. \#of genetic operations per rounds $\#\mathtt{Gen}$; Size of population pool $N_p$.}
   \State $\mathcal{P}= \{\}$.
   \State $\mathtt{DataOracle}\gets \mathtt{ConstructDataOracle}(\phi_{gt})$.
   \For{$i\gets 1 \text{ to } n$} \Comment{Initialize}
   \State $\mathbf{x}_c= \{x_1, \ldots, x_n\}\setminus \{x_i\}$.
   \State $\{\mathcal{D}_k\}_{k=1}^K\gets \mathtt{DataOracle}( \mathbf{x}_c, K)$.
   \State $\mathcal{P}_{new}\gets \mathtt{BuildGPPool}(\mathbf{x}_c,O_p, \{\mathcal{D}_k\}_{k=1}^K)$.
   \State $\mathcal{P}\gets \mathcal{P}\cup \mathcal{P}_{new}$.
   \EndFor

   \For{$i\gets 1 \text{ to } n$}
   \State $\mathcal{P}_{new}\gets \emptyset$.
        \For{$\langle \phi_{new}, \pi,\mathbf{x}_c \rangle \in \mathcal{P}$} \Comment{Control variable trials}
            \State $\{\mathcal{D}_k\}_{k=1}^K \gets \mathtt{DataOracle}(\mathbf{x}_c,K)$.
            \State $\mathbf{o}, \mathbf{c} \gets\mathtt{Optimize}(\phi_{new}, \{\mathcal{D}_k\}_{k=1}^K)$. 
            \State $\mathcal{P}_{new}\gets \mathcal{P}_{new}\cup \{\langle \phi_{new}, \mathbf{o}, \mathbf{c},\pi, \mathbf{x}_c\rangle\}$.
        \EndFor 
        \State $\mathcal{P}, \mathcal{H} \gets \mathtt{GP}(\mathcal{P}_{new}, \mathtt{DataOracle}, K, \#\mathtt{Gen},  O_p)$.
        \State $\mathcal{P}\gets {{\mathtt{TopK}}}(\mathcal{P}, K=N_p)$.  \Comment{Racing experiment schedule}
        \For{$\langle \phi, \pi,\mathbf{x}_c \rangle \in \mathcal{P}$} 
            \State $\phi\gets\mathtt{FreezeEquation}(\phi)$.
            
            \State randomly draw $x'\sim\mathbf{x}_c$.
            \State reduce control variables $\mathbf{x}_c\gets\mathbf{x}_c\setminus x'$. 
            \State update schedule  $\pi.\mathtt{append}(\mathbf{x}_c)$.
        \EndFor
    
   \EndFor
   % \EndFor
    \Return The set of hall-of-fame equations  $\mathcal{H}$.
\end{algorithmic}
\end{algorithm}

\paragraph{Running Time Analysis.}

The major hyper-parameters that impact the running time of \method are 1) the number of  genetic operations per round $M$; 2) total rounds $n$; 3) the maximum size of population pool $N_p$. A rough estimation of the time complexity of the proposed \method is $\mathcal{O}(nMN_p)$, which is the same as the CVGP algorithm. Another implicit factor of running time is the number of open constants $|\mathbf{c}|$ for every expression $\phi(\mathbf{x},\mathbf{c})$. An expression with more open constants needs more time for optimizers (like BFGS, CG) or more advanced optimizers (like Basin Hopping~\cite{wales1997global}) to find the solutions. We leave it to the empirical time evaluation in Table~\ref{tab:optimizer}.

\section{Related Work}

Early works in Symbolic regression are based on heuristic search~\cite{LANGLEY1981DataDiscovery,LENAT1977ubiquity}.
Genetic programming turns out to be effective in searching for good candidates of symbolic expressions~\cite{journal/2020/aifrynman,DBLP:conf/gecco/VirgolinAB19,DBLP:conf/gecco/HeLYLW22}. Reinforcement learning-based methods propose a risk-seeking policy gradient to find the expressions~\cite{DBLP:conf/iclr/PetersenLMSKK21,DBLP:conf/nips/MundhenkLGSFP21}. 
Other works use RL to adjust the probabilities of genetic operations~\cite{DBLP:journals/apin/ChenWG20}. Also, there are works that reduced the combinatorial search space by considering the composition of base functions, \textit{e.g.} Fast function extraction~\cite{mcconaghy2011ffx} and elite bases regression~\cite{DBLP:conf/icnc/ChenLJ17}.  
In terms of the families of expressions, research efforts  have been devoted to searching for polynomials with single or two variables~\cite{DBLP:journals/gpem/UyHOML11}, time series equations~\cite{DBLP:conf/icml/BalcanDSV18}, and also equations in physics~\cite{journal/2020/aifrynman}.

Multi-variable symbolic regression is more challenging since the search space increases exponentially with respect to the number of input variables. Existing works for multi-variable regression are mainly based on pre-trained encoder-decoder methods with a massive training dataset (e.g., millions of datasets~\cite{DBLP:conf/icml/BiggioBNLP21}), and even larger generative models (e.g., about 100 million parameters~\cite{DBLP:conf/nips/KamiennydLC22}).
Our \method is a tailored algorithm to solve multi-variable symbolic regression problems.

Our work is relevant to a line of work~\cite{DBLP:conf/ijcai/Langley77,DBLP:conf/ijcai/Langley79,DBLP:conf/ijcai/LangleyBS81,king2004functional,king2009autosci,cerrato2023rlsci} that implemented the human scientific discovery process using AI, pioneered by the BACON systems~\cite{DBLP:conf/ijcai/Langley77,DBLP:conf/ijcai/Langley79,DBLP:conf/ijcai/LangleyBS81}. 
While BACON's discovery was driven by rule-based engines and our CVGP uses modern machine-learning approaches such as genetic programming.

Choice of variables is an important topic in AI, including variable ordering for the construction of decision diagrams~\cite{DBLP:journals/informs/CappartBRPP22}, variable selection in tree search~\cite{DBLP:conf/nips/Song0H022}, variable elimination in probabilistic inference~\cite{DBLP:series/synthesis/2019Dechter,DBLP:conf/uai/DerkinderenHMKR20} and  backtracking search in solving constraint satisfaction problems~\cite{DBLP:journals/cin/Ortiz-BaylissAC18,DBLP:journals/heuristics/LiFY20,DBLP:journals/eaai/SongC0X022}.  Our method is one variant of variable ordering to the symbolic regression domain.  

Our work is also relevant to experiment design,  which studies the problem of drawing a minimum number of data for determining coefficients in linear regression models~\cite{10.1214/aos/1069362742,10.1214/12-AOS992,DBLP:journals/corr/abs-2301-08336}. Our work considers reducing the number of total data needed to uncover the ground truth expression.

\section{Experiments}
This section  demonstrates \method finds the symbolic expressions with the smallest Normalized Mean-Square Errors (NMSE)  (in Table~\ref{tab:trig-nmse} and Table~\ref{tab:livermore2-nmse}) and takes less computational time (in Figure~\ref{fig:quartile-time-partial}),
among all competing approaches on several noiseless datasets. 
In the ablation studies, we show our \method is consistently better than the baselines when evaluated in different evaluation metrics (in Figure~\ref{fig:different-metrics}). Also, our \method methods save a great portion of time than evoke CVGP with all the possible schedules. We present exact expressions in every dataset in appendix~\ref{apx:dataset-config}.

\subsection{Experimental Settings}
\noindent\textbf{Datasets.} We consider several public-available and multi-variable datasets, including 1) Trigonometric datasets~\cite{DBLP:arxiv/ecml/ny23}, 2) Livermore2 datasets~\cite{DBLP:conf/iclr/PetersenLMSKK21}, 3) Feynamn datasets~\cite{journal/2020/aifrynman}. For the consistency of presenting the experiments, each dataset is further partitioned by the number of variables for the expressions. 

\begin{table}[!t]
    \centering
    \caption{On trigonometric datasets, Median (50\%) and 75\%-quantile NMSE values of the expressions found by all the algorithms. 
    Our \method finds symbolic expressions with the smallest NMSEs.  ``$T.O.$'' implies the algorithm is timed out for 48 hours. The 3-tuples at the top $(\cdot,\cdot,\cdot)$ indicate the number of input variables, singular terms, and cross terms in the expression.}
    \label{tab:trig-nmse}
    \resizebox{\columnwidth}{!}{%
    \begin{tabular}{r|cc|cc|cc|cc|cc}
    \hline
      &\multicolumn{10}{c}{Datasets containing operators $O_p=\{\sin, \cos,+,-,\times\}$.} \\
         & \multicolumn{2}{c}{(3, 2, 2)} & \multicolumn{2}{c}{(4, 4, 6)} &\multicolumn{2}{c}{(5, 5, 5)}  &\multicolumn{2}{c}{(6, 6, 10)} & \multicolumn{2}{c}{(8, 8, 12)} \\
         & $50\%$&  $75\%$ & $50\%$&  $75\%$ & $50\%$&  $75\%$ & $50\%$&  $75\%$ & $50\%$&  $75\%$\\  \hline
        {\small \method(ours)} & $<\mathbf{1e-6}$ & $<\mathbf{1e-6}$ & $\mathbf{0.016}$ & $\mathbf{0.021}$ & $\mathbf{0.043}$ & $\mathbf{0.098}$  & $\mathbf{0.069}$ & $\mathbf{0.104}$ & $\mathbf{0.095}$ & $\mathbf{0.286}$\\ 
       CVGP & $0.039$ & $0.083$  & ${0.028}$ & $0.132$ &${0.086}$ & ${0.402}$ &  ${0.104}$&  ${0.177}$  & $T.O.$ &  $T.O.$ \\
       GP  & $0.043$ & $0.551$ & $0.044$ & ${0.106}$ & ${0.063}$ & ${0.232}$ &  $0.159$&  $0.230$ & $T.O.$ &  $T.O.$ \\
       DSR  & $0.227$ & $7.856$ & $2.815$ & $9.958$ & $2.558$ & $ 3.313$ &  $6.121$&  $16.32$ & $0.335$	& $0.410$ \\
       PQT & $0.855$ & $2.885$  & $2.381$ & $13.84$  & $2.168$ & $ 2.679$ &  $5.750$&  $16.29$ & $0.232$	 & $0.313$\\
        VPG  & $0.233$ & $0.400$ & $2.990$ & $11.32$ & $1.903$ & $ 2.780$ &  $3.857$&  $19.82$ & $0.451$	 & $0.529$\\
       GPMeld & $0.944$  & $1.263$ & $1.670$ & $2.697$ & $1.501$ & $2.295$ &  $7.393$&  $21.71$ & $T.O.$ &  $T.O.$ \\
       Eureqa & $<\mathbf{1e-6}$ & $<\mathbf{1e-6}$ & $0.024$ & $0.122$ & $0.158$ & $0.377$ &  $0.910$&  $1.927$ & $0.162$ & $2.223$\\
        \hline 
    \end{tabular}%
    }
    % \vspace{-1em}
\end{table}

\noindent\textbf{Evaluation Metrics.} We mainly consider two evaluation criteria for the learning algorithms tested in our work: 1) The goodness-of-fit measure (NMSE), indicates how well the learning algorithms perform in discovering symbolic expressions. The median (50\%) and 75\%-quantile of the NMSE are reported. 
2) the total running time of each learning algorithm.  The computation of the total running time encompasses the duration taken for each program to uncover a promising expression. It's worth noting that this calculation incorporates the time spent on data oracle queries. Additionally, we've implemented a strict time limit of 48 hours to prevent programs from exceeding reasonable execution times.

Given a testing dataset $\mathcal{D}_{\text{test}}=\{(\mathbf{x}_{i},y_i)\}_{i=1}^n$ generated from the ground-truth expression $\phi$, we measure the goodness-of-fit of a predicted expression $\bar{\phi}$, by evaluating the mean-squared-error (MSE), normalized-mean-squared-error (NMSE), root mean-square
error (RMSE), normalized root Mean-squared error (NRMSE):
\begin{equation}\label{eq:loss-function}
\begin{aligned}
\text{MSE}&=\frac{1}{n}\sum_{i=1}^n(y_{i}-\bar{\phi}(\mathbf{x}_{i}))^2, \qquad\quad \text{NMSE}=\frac{\frac{1}{n}\sum_{i=1}^n(y_{i}-\bar{\phi}(\mathbf{x}_{i}))^2}{\sigma_y^2} \\
\text{RMSE}&=\sqrt{\frac{1}{n}\sum_{i=1}^n(y_{i}-\bar{\phi}(\mathbf{x}_{i}))^2} ,\qquad
\text{NRMSE}=\frac{1}{\sigma_y} \sqrt{\frac{1}{n}\sum_{i=1}^n(y_{i}-\bar{\phi}(\mathbf{x}_{i}))^2} 
\end{aligned}
\end{equation}
where the empirical variance $\sigma_y=\sqrt{\frac{1}{n}\sum_{i=1}^n \left(y_i-\frac{1}{n}\sum_{i=1}^n y_i\right)^2}$. Note that the coefficient of determination ($R^2$) metric~\cite{nagelkerke1991note,DBLP:conf/nips/CavaOBFVJKM21} is equal to $(1-\text{NMSE})$ and therefore omitted in the experiments. 

For the fairness of the evaluation, every learning algorithm outputs the most probable symbolic expression. We then apply the same testing set $\mathcal{D}_{\text{test}}$ to compute the goodness measure in Equation~\ref{eq:loss-function} and report the median (50\%) and 75\% quartile values instead of mean value over all the expressions in that group of dataset. This is a common practice of reporting results for combinatorial algorithms, to avoid the performance comparison heavily impacted by outliers. The outliers in symbolic regression are those expressions very challenging to discover so the learning algorithm can only predict sub-optimal expressions with large NMSE loss values.

\noindent\textbf{Baselines.}  We consider the following baselines based on evolutionary algorithms: 1) Genetic Programming (GP)~\cite{DEAP_JMLR2012}.  2) Eureqa~\cite{DBLP:journals/gpem/Dubcakova11}, which is the current best commercial software based on evolutionary search algorithms. 
We also consider a series of baselines using reinforcement learning: 3) Priority queue training (PQT)~\cite{DBLP:journals/corr/abs-1801-03526}. 4) Vanilla Policy Gradient (VPG) that uses the REINFORCE algorithm~\cite{DBLP:journals/ml/Williams92} to train the model. 5) Deep Symbolic Regression (DSR)~\cite{DBLP:conf/iclr/PetersenLMSKK21}. 6) Neural-Guided Genetic Programming Population Seeding (GPMeld)~\cite{DBLP:conf/nips/MundhenkLGSFP21}. Note that recent work Symbolic Physics Learner~\cite{DBLP:conf/iclr/Sun0W023} does not support solving for open constants. SPL can only work for expressions with no constants. Thus it is omitted.

\paragraph{Hyper-parameter Configuration} We leave detailed descriptions of the configurations in  Appendix~\ref{apx:hyper-parameter-config} of our \method and baseline algorithms in Appendix B and only mention a few implementation notes here. 
Our \method uses a data oracle, which returns (noisy) observations of the ground-truth equation when queried with inputs. 
We cannot implement the same Oracle for other baselines because of code complexity and/or no available code. 
To ensure fairness, the sizes of the training datasets we use for those baselines are larger than the total number of data points accessed in the full execution of those algorithms. 
In other words, their access to data would have no difference if the same oracle has been implemented for them because it does not affect the executions whether the data is generated ahead of the execution or on the fly. 
The reported NMSE scores in all charts and tables are based on separately generated data that have never been used in training.  
The threshold to freeze operators in \method is if the MSE to fit a data batch is below $1e-3$. The threshold to freeze the value of a constant in \method is if the variance of best-fitted values of the constant across trials drops below $0.001$.

\subsection{Experimental Result Analysis}

\begin{table*}[!ht]
    \centering
    \caption{On Livermore2 and Feynman datasets, Median (50\%) and 75\%-quantile NMSE values of the symbolic expressions found by all the algorithms. 
    Our \method finds symbolic expressions with the smallest NMSEs. $n$ is the number of independent variables in the expressions.}
    \label{tab:livermore2-nmse}
    \resizebox{\columnwidth}{!}{%
    \begin{tabular}{r|cc|cc|cc|cc|cc}
    \hline
    &\multicolumn{6}{c|}{Livermore2} &\multicolumn{4}{c}{Feynman } \\
    
         & \multicolumn{2}{c}{$n=4$} &\multicolumn{2}{c}{$n=5$} &\multicolumn{2}{c|}{$n=6$}  & \multicolumn{2}{c}{$n=4$} &\multicolumn{2}{c}{$n=5$}  \\
         & $50\%$&  $75\%$ & $50\%$&  $75\%$  & $50\%$&  $75\%$ & $50\%$&  $75\%$& $50\%$&  $75\%$\\ \midrule
        {\small \method(ours)} &  $<\mathbf{1e\mbox{-}6}$ & $\mathbf{2.03e\mbox{-}3}$ & $\mathbf{0.004}$ & $\mathbf{0.047}$ & $\mathbf{0.001}$ & $\mathbf{0.073}$ & $\mathbf{0.0154}$ & $\mathbf{0.195}$ & ${0.577}$ & $\mathbf{0.790}$\\
        CVGP  & $0.052$ & $0.810$ & $0.275$ & $1.007$ & $0.328$ & $1.012$ & $1.002$ & $1.010$ & $1.001$ & $1.002$ \\
        GP &   $0.059$ & $0.962$ & $0.331$ & $1.003$ & $1.001$ & $1.026$ & $1.003$ & $1.010$ & $1.002$ & $1.011$ \\
        DSR &  $0.0301 $ & ${0.0483}$ & $0.050$ & $0.284$ & $0.230$ & $0.486$ & $0.216$ & $0.920$ & $0.976$ & $1.001$\\
        PQT &  $0.042 $ & $0.063$ & $0.074$ & $0.227$ & $0.170$ & $0.410$ & $0.172$ & $0.765$& $1.003$ & $1.027$\\
        VPG &  $0.0368 $ & $0.0737$ & $0.093$ & $0.322$ & $0.206$ & $0.535$ & $0.188$ & $0.971$ & $1.006$ & $1.025$\\
        GPMeld & $0.029$ & $0.061$ & $0.049$ & $0.259$ & $0.144$ &  $0.504$ & $0.177$ & $0.708$ & $0.940$ & $1.002$\\
        Eureqa & $0.508$ & $0.980$ & $0.083$  & $0.249$  & $0.026$  & $0.302$  & $0.026$  & $0.397$  & $\mathbf{0.434}$ & $0.943$ \\
        \hline
    \end{tabular}%
    }
\end{table*}

\noindent\textbf{Goodness-of-fit Benchmark}
Our \method attains the smallest median (50\%) and 75\%-quantile NMSE values among all the baselines when evaluated on selected Trigonometric, Livermore2, and Feynman datasets (Table~\ref{tab:trig-nmse}). This shows our method can better handle multiple variables symbolic regression problems than the current best algorithms in this area.  For the Trigonometric dataset with $n=8$ variables, both GP and CVGP take more than 2 days to find the optimal expression. The reason is that there are too many open constants in each expression in the population pool, making the optimization problem itself non-convex problems and time-consuming to find the solution. This behavior is another indication that CVGP is stuck at some unfavorable experiment schedule.

\begin{figure}[!t]
    \centering
      \includegraphics[width=0.355\linewidth]{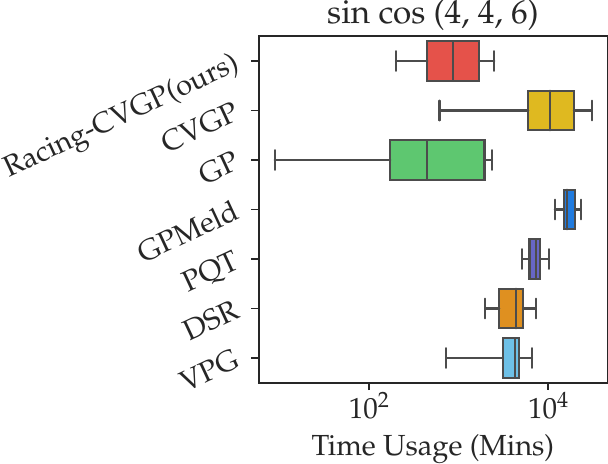}
     \includegraphics[width=0.205\linewidth]{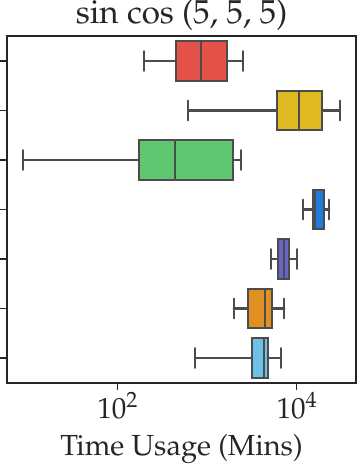}
    \caption{On selected Trigonometric datasets, quartiles of the total running time of all the methods. Our \method method takes less time than CVGP by early stopping those unfavorable experiment schedules.}
    \label{fig:quartile-time-partial}
\end{figure}

\noindent\textbf{Empirical Running Time Analysis.}
We summarize the running time analysis in Figure~\ref{fig:quartile-time-partial}. Our \method method takes less time than CVGP as well as the rest baselines. The main reason is early stop those unfavorable experiment schedules. %See Equation~\ref{eq:loss-function} for more figures.

\begin{figure}[!t]
    \centering
      \includegraphics[width=0.33\linewidth]{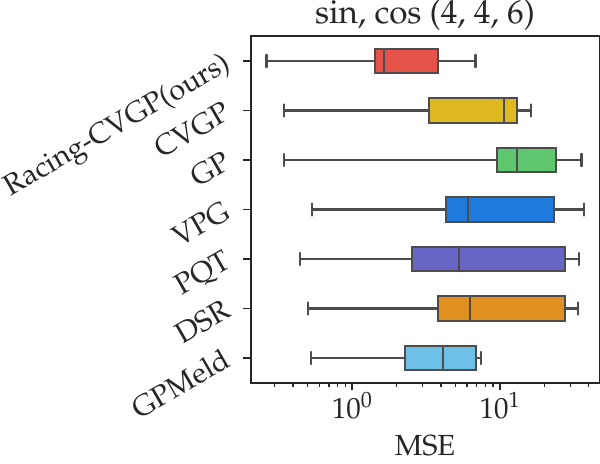}
    \includegraphics[width=0.20\linewidth]{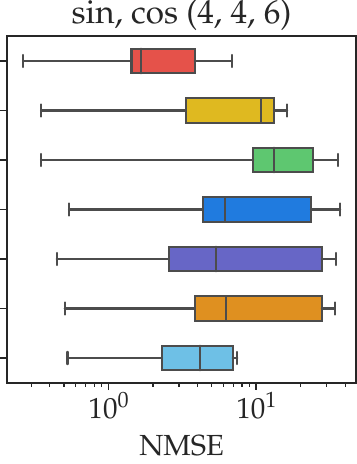}
    \includegraphics[width=0.20\linewidth]{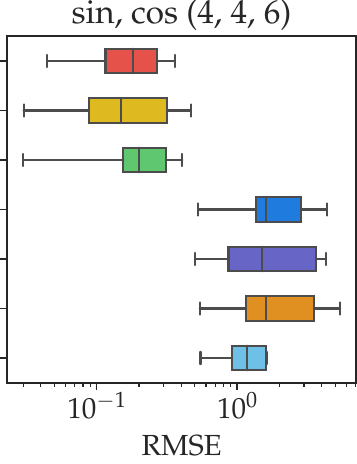} 
    \includegraphics[width=0.20\linewidth]{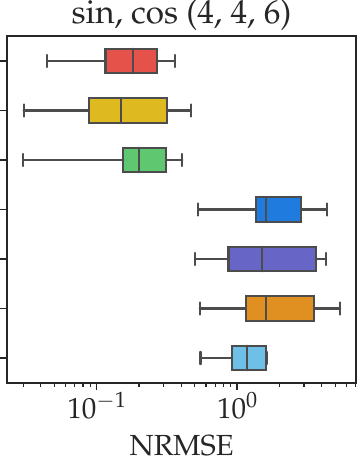}
    \caption{On selected Trigonometric datasets, MSE, NMSE, RMSE, and NRMSE  evaluation metrics of the expressions found by different algorithms.}
    \label{fig:different-metrics}
\end{figure}

\subsection{Ablation Studies} \label{sec:ablation}

\begin{figure}
    \centering
    \includegraphics[width=0.43\linewidth]{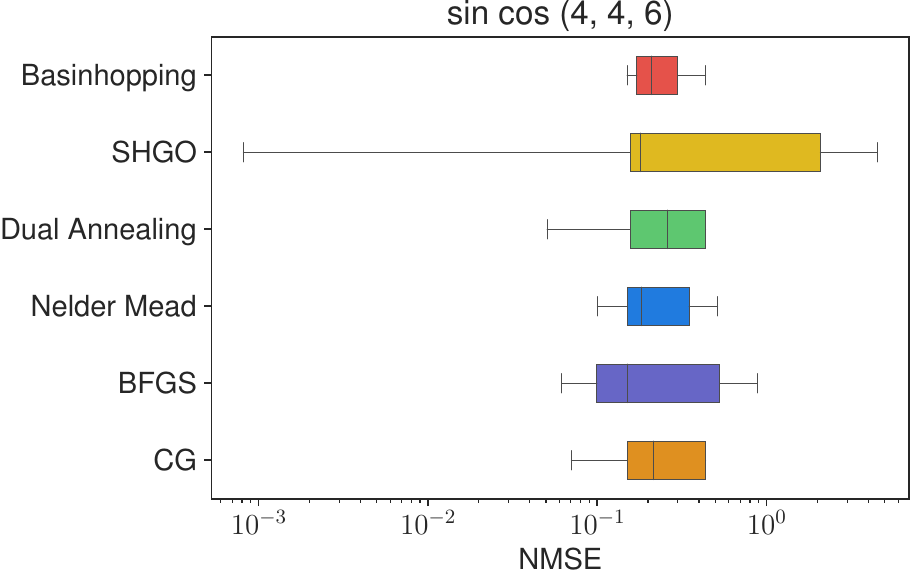}
    \hfill
    \includegraphics[width=0.43\linewidth]{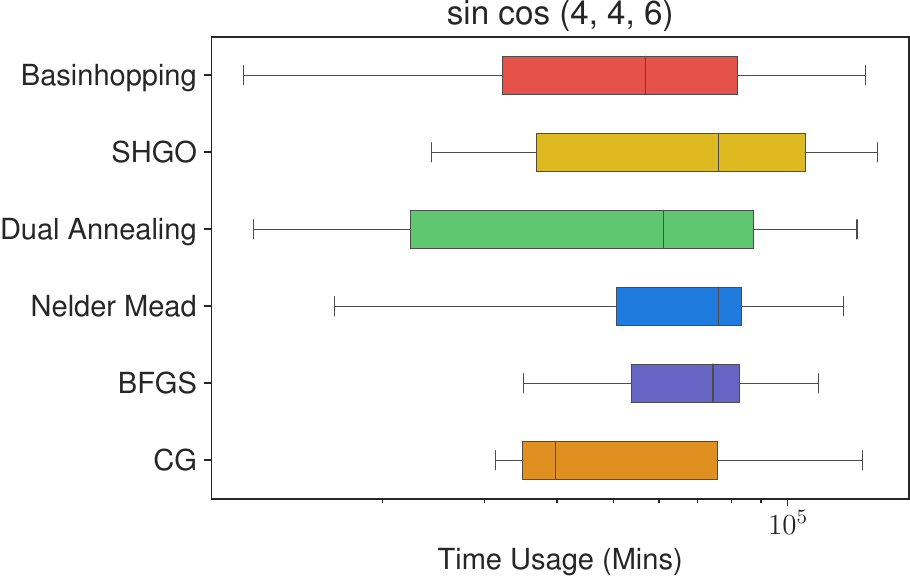}
    \caption{Impact of optimizers on finding the values of open constants for non-convex expressions. Over 10 randomly generated expressions involving 4 variables,  SHGO can find better solutions (in terms of NMSE metric) than local optimizers (including Nelder-Mead, BFGS, CG), while the time taken by SHGO is higher than local optimizers. }
    \label{tab:optimizer}
\end{figure}

\noindent\textbf{Impact of Optimizers.} 
Here we study the impact of using global and local optimizers over those non-convex expressions. With the introduction of control variable experiments, fitting the open constants in the expressions is solving more and more non-convex optimization problems.  We consider several optimizers:  CG~\cite{fletcher1964function} Nelder-Mead~\cite{DBLP:journals/coap/GaoH12}, BFGS~\cite{fletcher2000practical}, Basin Hopping~\cite{wales1997global}, SHGO~\cite{DBLP:journals/jgo/EndresSF18}, Dual Annealing~\cite{tsallis1996generalized}.
The list of local and global optimizers shown in Figure~\ref{tab:optimizer} are from Scipy library\footnote{\url{https://docs.scipy.org/doc/scipy/reference/optimize.html}}.

For those expressions in the populations, an optimizer might find a set of open constants for a structurally correct expression with large NMSE errors, resulting in a low ranking in the whole population. Such structurally correct expressions will not be included after several rounds of genetic operations.

We summarize the experimental result in Figure~\ref{tab:optimizer}. In general, the list of global optimizers (SHGO, Direct, Basin-Hopping, and  Dual-Annealing) fits better for the open constants than the list of local optimizers but they take significantly more CPU resources and time for computations.

\paragraph{Impact of Fitness Measure.} We collect the benchmark of different evaluation metrics in Figure~\ref{fig:different-metrics}, \textit{i.e.,} MSE, NMSE, during testing over the selected Trigonometric datasets. The RMSE and NRMSE evaluation metrics are available in Equation~\eqref{eq:loss-function}.

\begin{figure}[!t]
    \centering
     \includegraphics[width=0.5\linewidth]{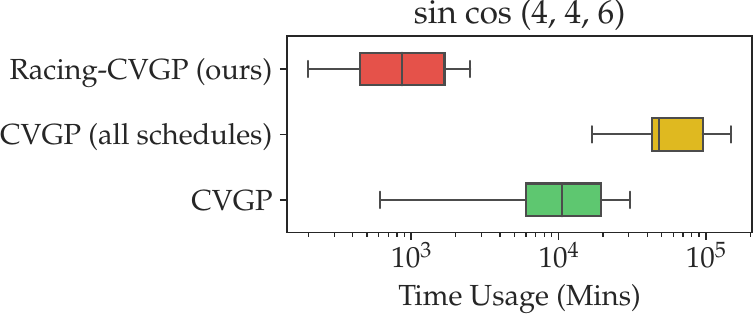}
    \caption{On selected Trigonometric datasets, quartiles of the total running time of \method, CVGP, and CVGP with all the experiment schedules. Our \method saves a great portion of time compared with CVGP with all the schedules for expressions with $n=4$ variables.}
    \label{fig:quartile-time-exp-schedule}
\end{figure}

\paragraph{Time-Saving Analysis.} We further collect the time comparison between our \method and the CVGP (with all the experiment schedules) in Figure~\ref{fig:quartile-time-exp-schedule}.
The quartiles of time distribution over $10$ random expressions with $4$ variables show that  Our \method saves a great portion of time compared with \textit{CVGP with all the schedules}.

Note that we implement the CVGP (with all the experiment schedules) with a tree structure to maintain all schedules. The exact implementation can be found at Appendix~\ref{apx:baseline} and also in Figure~\ref{fig:all-cvgp}.

\begin{figure}[!t]
\centering
\includegraphics[width=0.49\linewidth]{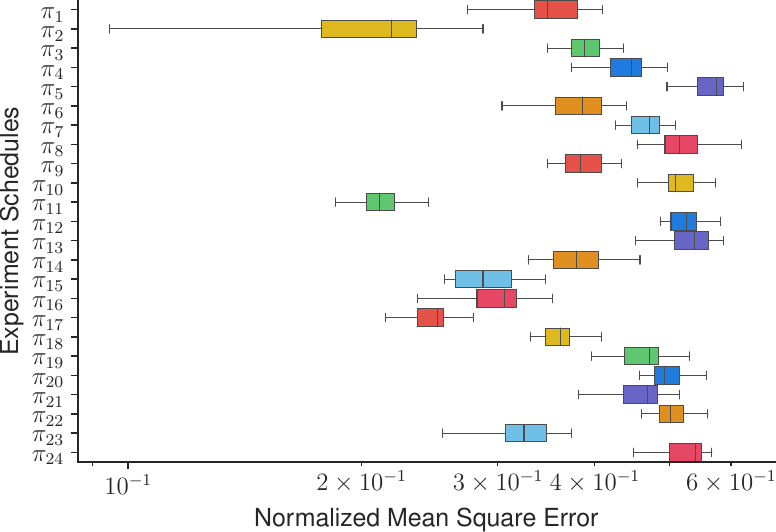}
\includegraphics[width=0.49\linewidth]{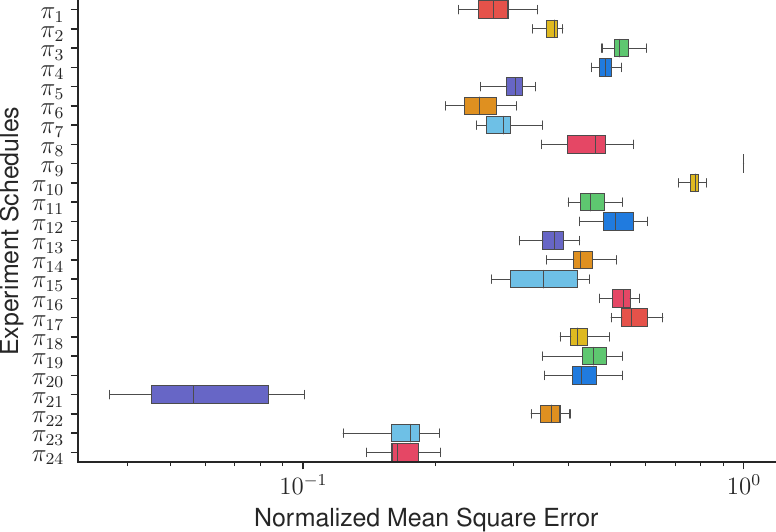}
\includegraphics[width=0.49\linewidth]{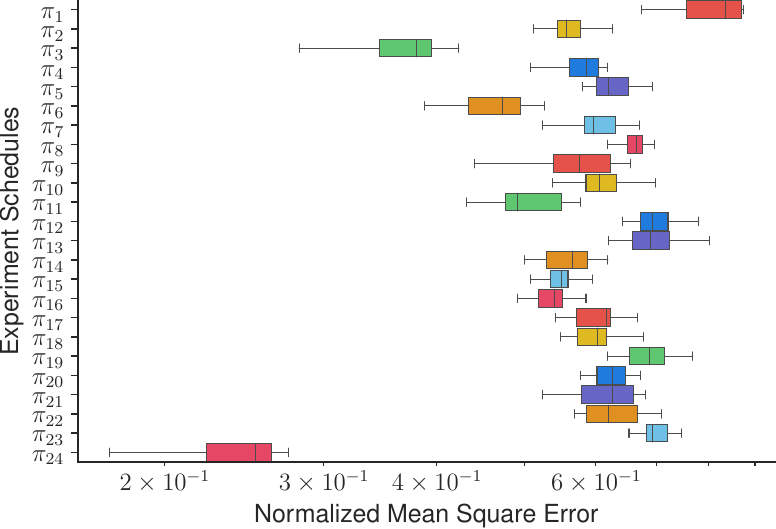}
\includegraphics[width=0.49\linewidth]{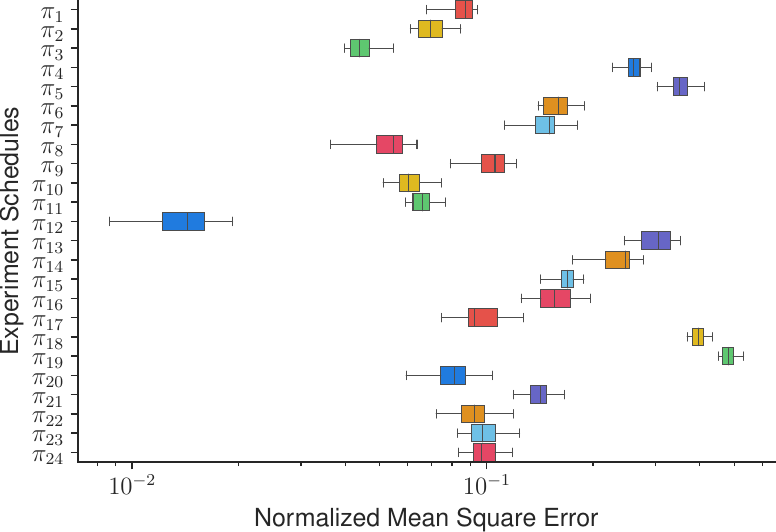}
\includegraphics[width=0.49\linewidth]{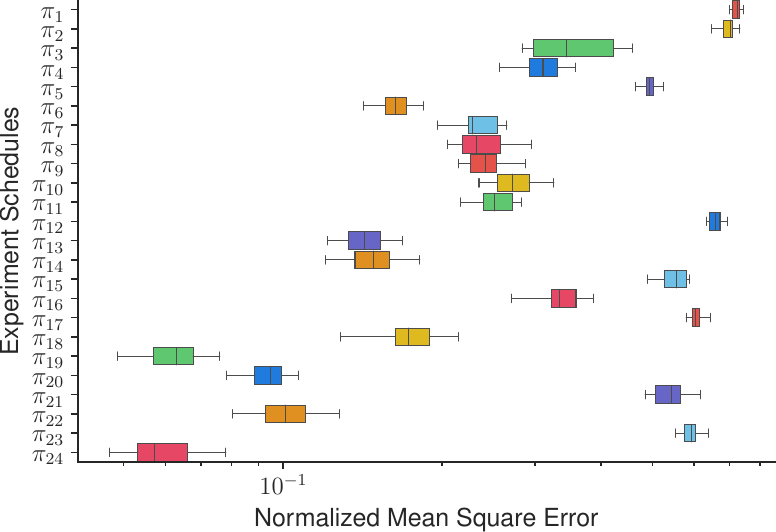}
\includegraphics[width=0.49\linewidth]{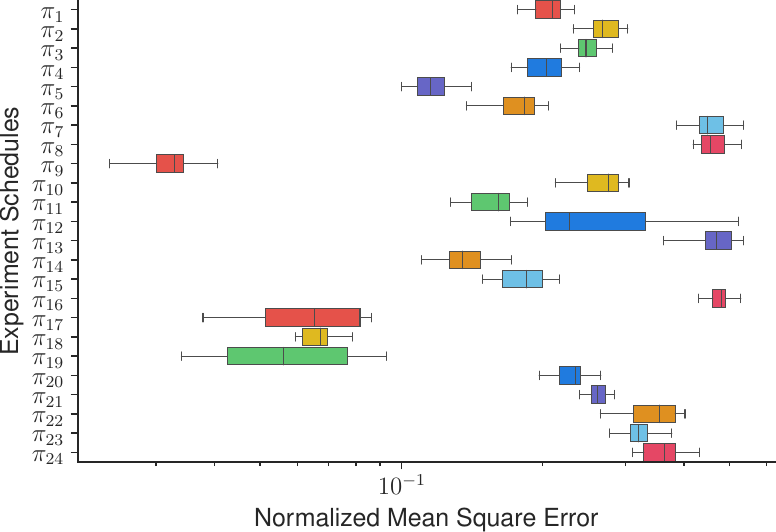}
\caption{Impact of experiment schedules (noted as $\pi$) on learning performance of control variable genetic programming, on the Trigonometric 
 $(4,4,6)$ with operator set $\{+,-,\times, \div, \sin,\cos\}$ dataset. For the discovery of 10 different expressions with $4$ variables, there always exists a better experiment schedule than the default one (\textit{i.e.}, $\pi_1$), in terms of normalized mean square error.}
\label{fig:apx-expriment-schedule-1}
\end{figure}

\paragraph{Impact of Experiment Schedules} In Figures~\ref{fig:apx-expriment-schedule-1} and~\ref{fig:apx-expriment-schedule-2}, we summarize the result of running the same CVGP algorithm with different experiment schedules, on the Trigonometric $(4,4,6)$ with operator set $\{+,-,\times, \div, \sin,\cos\}$ dataset.
For the discovery of 10 different expressions with $4$ variables, 1) there always exists a better experiment schedule than the default one (\textit{i.e.}, $\pi_1$), in terms of normalized mean square error. 2) The performance of the same CVGP algorithm varies greatly with different experiment schedules.

\begin{figure}[!ht]
\centering
\includegraphics[width=0.49\linewidth]{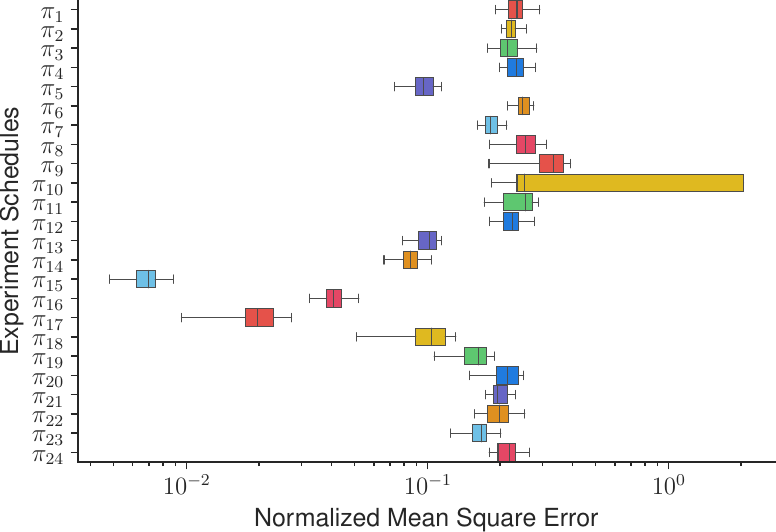}
\includegraphics[width=0.49\linewidth]{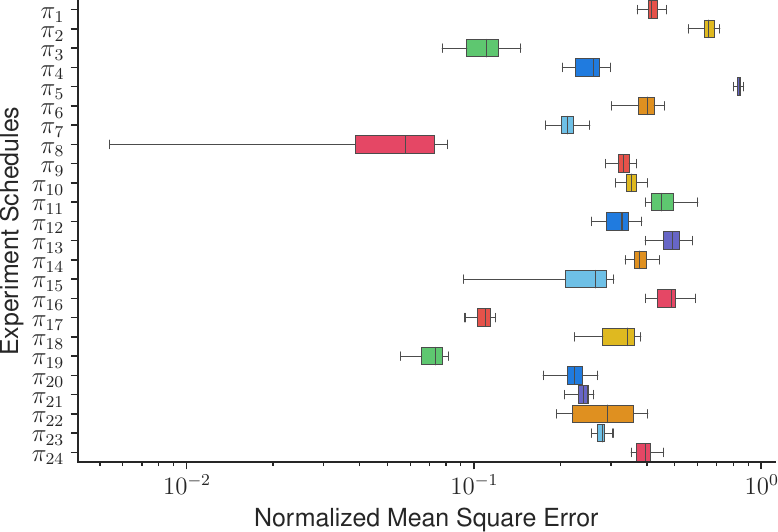}
\includegraphics[width=0.49\linewidth]{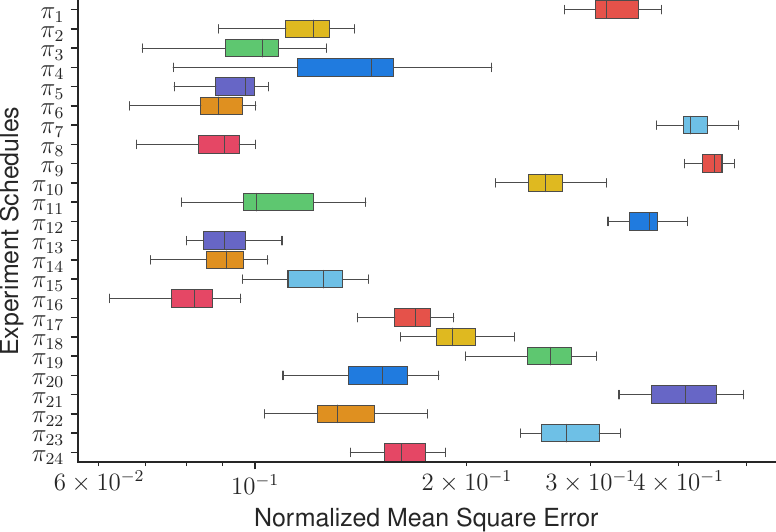}
\includegraphics[width=0.49\linewidth]{exp/ordering/inv4469.neg_nmse.order.pdf}
\caption{(Continued) Impact of experiment schedules (noted as $\pi$) on learning performance of control variable genetic programming. For the discovery of expression with $4$ variables, there always exists a better experiment schedule than the default one (\textit{i.e.}, $\pi_1$), in terms of normalized mean square error.}
\label{fig:apx-expriment-schedule-2}
\end{figure}

\section{Conclusion}
In this research, we propose Control Variable Genetic Programming (\method) for symbolic regression with many independent variables. Our \method can accelerate the regression process by discovering equations from promising experiment schedules and early stop those unfavorable experiment schedules. 
We evaluate Racing-CVGP on several synthetic and real-world datasets corresponding to true physics laws. 
We demonstrate that Racing-CVGP outperforms CVGP and 
a series of symbolic regressors which discover equations from fixed datasets.

\section*{Acknowledgments}
This research was supported by
NSF grants IIS-1850243, CCF-1918327.

\clearpage
\bibliography{reference}
\bibliographystyle{unsrtnat} 
\newpage

\appendix
\newpage

% \section{Implementation} \label{apx:implement}

% Please find our code repository.

% It contains  1) the implementation of our \method method, 2) the list of datasets, and 3) the implementation of several baseline algorithms.

\section{Genetic Programming Algorithm in \method} \label{apx:gp}

\begin{figure}[!t]
    \centering
    \includegraphics[width=1\linewidth]{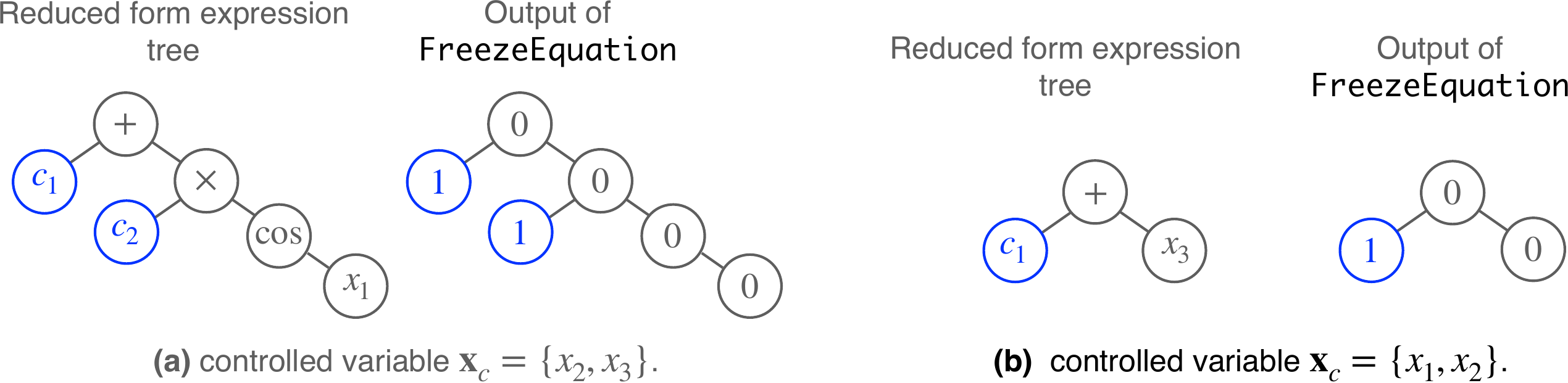}
    \caption{Visualization of the $\mathtt{FreezeEquation}$ function. The value ``$0$'' in the code implies the corresponding node is non-editable and $1$ implies the corresponding node is editable, by the GP algorithm.  The $\mathtt{FreezeEquation}$ function will increase the probability of finding expression trees with close-to-zero fitness scores.}
    \label{fig:freeze}
\end{figure}

For the $\mathtt{FreezeEquation}$ function used in Algorithm~\ref{alg:racing-cvgp}, we use Figure~\ref{fig:freeze} to demonstrate the output. The $\mathtt{FreezeEquation}$ function will reduce the size of candidate nodes to be edited in the GP algorithms and increase the probability of finding expression trees with close-to-zero fitness scores.

The execution of $\mathtt{GP}$ function in our \method framework is presented in Algorithm~\ref{alg:gp}.  It is a minimally modified genetic programming algorithm for symbolic regression.

For the $\mathtt{Mutate}$ step, the algorithm will apply one of the following operations over the input expression tree and output a modified expression tree.
\begin{enumerate}
    \item Find a leaf node that is not frozen and then replace the node with a generate a full expression tree of maximum depth involving variables only in $\mathbf{x}\setminus\mathbf{x}_c$.
    \item Find a node and replace it with a node of the same arity. Here arity is the number of operands taken by an operator.  For example, the arity of binary operators $\{+,-,\times,\div\}$ is $2$ and the arity of unary operators $\{\sin,\cos ,\log,\exp\}$ is $1$.
    \item Inserts a node at a random position, and the original subtree at the location becomes one of its subtrees.
    \item Delete a node that is not frozen, use one of its children to replace its position.
\end{enumerate}

For the $\mathtt{Mate}$ step, we will pick two expressions $\phi_{l},\phi_{j}$ from the population pool $\mathcal{P}$ that has the same control variables $\mathbf{x}_{c,l}=\mathbf{x}_{c,j}$. Then we exchange two randomly chosen subtrees in the expressions. Because applying mating over two expressions with different control variables does not necessarily result in two better expressions. 

\begin{algorithm}[!t]
   \caption{$\mathtt{GP}(\mathcal{P}, \mathtt{DataOracle}, K, M, \mathtt{\#Gen}$, $\mathtt{\#Hof}$, $P_{mu}$, $P_{ma}$, $O_p)$}
   \label{alg:gp}
   \begin{algorithmic}[1]
   \Require{ $\mathtt{DataOracle}$; \# control variable trials $K$; mutation node library $O_p$; GP Pool $\mathcal{P}$;  GP pool size $N_p$; \#generations $\mathtt{\#Gen}$; \#expressions in hall-of-fame set $\mathtt{\#Hof}$; mutate probability $P_{mu}$; mate probability $P_{ma}$}. 
   \For{$i \gets 1 \textit{ to } \mathtt{\#Gen} $}
   \State $\mathcal{P}_{new} \gets \emptyset$;
   \For{$\langle \phi_{new}, \pi,\mathbf{x}_c\rangle \in \mathcal{P}$}
    \If{with probability $P_{mu}$}\Comment{{mutation}}
        \State $\phi_{new} \gets \mathtt{Mutate}(\phi_{new}, O_p, \mathbf{x}\setminus\mathbf{x}_c)$;
        \State $\{\mathcal{D}_k\}_{k=1}^K\gets\mathtt{DataOracle}( \mathbf{x}_c, K)$;
       \State $\mathbf{o}$, $\mathbf{c}$ $\gets \mathtt{Optimize}(\phi_{new},  \{\mathcal{D}_k\}_{k=1}^K)$;
    \EndIf
    \State $\mathcal{P}_{new}  \gets \mathcal{P}_{new}  \cup \{\langle \phi_{new}, \mathbf{o}, \mathbf{c},\pi, \mathbf{x}_c\rangle\}$;
   \EndFor
   \State $\mathcal{P}\leftarrow   \mathcal{P}_{new}$; $\mathcal{P}_{new}\leftarrow \emptyset$ 
   \For{$\langle \phi_{l}, \pi_l,\mathbf{x}_{c,l}\rangle, \langle \phi_{j}, \pi_{j},\mathbf{x}_{c,j}\rangle \in \mathcal{P}$} \Comment{mating}
    \If{ with probability $P_{ma}\; \text{ and }\;\mathbf{x}_{c,l}=\mathbf{x}_{c,j}$}  \Comment{pick two expressions with the same $\mathbf{x}_c$} 
    \State ${\phi_l, \phi_{j}} \gets \mathtt{Mate}(\phi_l, \phi_{j})$; 
    \State $\{\mathcal{D}_k\}_{k=1}^K\gets\mathtt{DataOracle}(\mathbf{x}_{c,l},K)$;
    \State $\mathbf{o}_l, \mathbf{c}_l$ $\gets \mathtt{Optimize}(\phi_{l},  \{\mathcal{D}_k\}_{k=1}^K)$;
    \State $\{\mathcal{D}_k\}_{k=1}^K\gets\mathtt{DataOracle}( \mathbf{x}_{c,j},K)$;
    \State $\mathbf{o}_j, \mathbf{c}_j$ $\gets \mathtt{Optimize}(\phi_{j}, \{\mathcal{D}_k\}_{k=1}^K)$;
    \EndIf
    \State $\mathcal{P}_{new} \gets \mathcal{P}_{new} \cup \{\langle \phi_{l}, \mathbf{o}_l, \mathbf{c}_l,\pi_l, \mathbf{x}_{c,l}\rangle, \langle \phi_{j}, \mathbf{o}, \mathbf{c}_j,\pi_j, \mathbf{x}_{c,j}\rangle\}$;
   \EndFor
   
   \State $\mathcal{H}\gets {{\mathtt{TopK}}}(\mathcal{P}_{new} \cup \mathcal{H}, K=\mathtt{\#Hof})$; \Comment{Update the hall of fame set}
   \EndFor
  \Return  GP pool and hall-of-fame $\mathcal{P}_{new}, \mathcal{H}$.
\end{algorithmic}
\end{algorithm}

\newpage
\section{Experiment Settings}\label{apx:exp-set}

\subsection{Dataset Configuration}\label{apx:dataset-config}
\paragraph{Trigometric Datasets.} We list all series of datasets in Tables~\ref{tab:sincos-332},~\ref{tab:sincos-446},~\ref{tab:sincos-555},~\ref{tab:sincos-6610},~\ref{tab:sincos-8812},~\ref{tab:sincos-8812ct}. 

\paragraph{Livermore2 Dataset.}
The list of Livermore2  datasets is modified from \footnote{\url{https://github.com/brendenpetersen/deep-symbolic-optimization/blob/master/dso/dso/task/regression/benchmarks.csv}}.
The reason for modification is that some expressions can easily output \texttt{Not-a-Number} or \texttt{Inf}, where all the learning algorithms cannot properly handle these outlier value cases.
In Tables~\ref{tab:livermore2-1}, \ref{tab:livermore2-2}, \ref{tab:livermore2-3}, we details the exact equation of Livermore2~\cite{DBLP:conf/iclr/PetersenLMSKK21}. The operator set for each expression is available in the codebase.

The list of Feynman datasets is collected from\footnote{\url{https://github.com/omron-sinicx/srsd-benchmark/blob/main/datasets/feynman.py}}. In Tables~\ref{tab:Feynman-4}, \ref{tab:Feynman-5}. We only use a subset of the expressions in the original Feynman dataset. The challenging part for this dataset is the ranges of input variables vary greatly. For example, in one equation with ID ``ICh34Eq8'' the ranges of all the variables are:
\begin{equation}
x_1\in (10^{-11}, 10^{-9}),x_2 \in (10^5, 10^7), x_3\in (10, 10^3), x_4\in (10^9, 10^{11})
\end{equation}
In comparison, the input ranges of Livermore2 dataset are $x_i\in(0.01, 10)$.
 The operator set for each expression is available in the codebase.

\begin{table}[!t]
    \centering
    \caption{Detailed expressions in Trigonometric datasets $(3,2,2)$ containing operators $\{\sin, \cos,+,-,\times\}$.}
    \label{tab:sincos-332}
    \begin{tabular}{c|cl}
    \toprule
        & \multicolumn{2}{c}{$\sin \cos (3, 3, 2)$} \\  
        Eq. ID & $n$ & Expression \\ \hline
 1 & $3$ & $0.6098x_1\sin(x_0) + 0.66x_2 - 0.5542\sin(x_2)\cos(x_1) - 0.5932\cos(x_0) + 0.1835$ \\
 2 & $3$ & $0.9272x_0\cos(x_1) - 0.8311x_0 - 0.7951x_1 + 0.5114\cos(x_1)\cos(x_2) - 0.8436$ \\
 3 & $3$ & $-0.0951x_0x_2 + 0.0127x_2\sin(x_1) - 0.5768x_2 - 0.2143\cos(x_0) - 0.6254$ \\
 4 & $3$ & $-0.3162x_0x_2 - 0.6406x_1x_2 - 0.802x_1 + 0.3979\cos(x_0) + 0.0068$ \\
 5 & $3$ & $0.7774x_0 - 0.5646x_1\sin(x_0) - 0.8781x_2 + 0.7823\sin(x_2)\cos(x_1) + 0.4612$ \\
 6 & $3$ & $-0.0999x_0\sin(x_1) - 0.4304x_0\cos(x_2) + 0.5153x_1 - 0.6365\cos(x_0) - 0.1823$ \\
 7 & $3$ & $0.6162x_0x_1 - 0.8577x_2\sin(x_0) - 0.8295x_2 + 0.3185\sin(x_1) - 0.0956$ \\
 8 & $3$ & $0.7621x_0x_1 - 0.5348x_1 - 0.8292x_2 + 0.4458\sin(x_2)\cos(x_1) + 0.2351$ \\
 9 & $3$ & $0.4681x_0 + 0.4856x_1x_2 - 0.8895x_2\sin(x_0) - 0.6741\cos(x_1) - 0.8204$ \\
 10 & $3$ & $-0.4634x_0\sin(x_2) - 0.7682x_2 - 0.4991\sin(x_1)\cos(x_2) + 0.1834\sin(x_1) + 0.3475$ \\
\hline
    \end{tabular}
\end{table}

\begin{table}[!t]
    \centering
    \caption{ Detailed expressions in Trigonometric datasets $(4,4,6)$ containing operators $\{\sin, \cos,+,-,\times\}$.}
    \label{tab:sincos-446}
    % \resizebox{\columnwidth}{!}{%
    \begin{tabular}{c|cl}
    \toprule
        & \multicolumn{2}{c}{$\sin \cos (4, 4, 6)$} \\  
        Eq. ID & $n$ & Expression \\ \hline
1 & $4$ & $0.0424x_1x_2 - 0.7582x_1 + 0.9181x_2x_3 - 0.587x_2\cos(x_0) + 0.2988x_2 - 0.9579x_3$\\
&& $+ 0.2076\sin(x_0)\cos(x_1) + 0.0865\sin(x_0) + 0.9965\sin(x_1)\cos(x_3)  $ \\
 && $+ 0.8622\cos(x_0)\cos(x_3) + 0.124$ \\ \hline
 2 & $4$ & $0.5998x_0x_1 + 0.5148x_0x_2 + 0.0606x_0x_3 + 0.1105x_1x_3 - 0.8742x_1 - 0.8527x_2x_3$ \\
 && $- 0.0896x_2\sin(x_1) + 0.2811x_2 + 0.8264\sin(x_0) + 0.0406\sin(x_3) + 0.4854$ \\\hline
 3 & $4$ & $-0.9296x_0 + 0.6272x_1\sin(x_0) + 0.4468x_2x_3 + 0.7135\sin(x_0)\cos(x_3) + 0.6816\sin(x_2)$ \\
 && $- 0.9374\sin(x_3)\cos(x_1) - 0.5579\sin(x_3) - 0.5481\cos(x_0)\cos(x_2) $ \\
 && $- 0.837\cos(x_1)\cos(x_2) - 0.3081\cos(x_1) - 0.1092$ \\ \hline
 4 & $4$ & $-0.802x_0x_1 - 0.4736x_0x_2 + 0.8366x_0\sin(x_3) - 0.7204x_1\cos(x_2) + 0.5086x_2x_3 -$\\
 && $ 0.9419x_2 - 0.8707x_3\cos(x_1) + 0.5934\sin(x_0) - 0.1084\sin(x_1) + 0.6729\sin(x_3) + 0.0363$ \\\hline
 5 & $4$ & $0.3847x_0x_3 - 0.904x_1\sin(x_0) - 0.3458x_1\sin(x_2) + 0.2652x_1\cos(x_3) + 0.9379x_1$\\
 &&$- 0.0158x_2\cos(x_0) - 0.0119x_2 - 0.6445x_3\sin(x_2) - 0.7881x_3 + 0.1602\sin(x_0) + 0.0368$ \\\hline
 6 & $4$ & $0.1068x_0\cos(x_1) - 0.9693x_0\cos(x_2) + 0.7863x_1x_3 - 0.8555x_1 - 0.2549x_3\sin(x_0)$ \\
 && $+ 0.3453\sin(x_0) + 0.2202\sin(x_1)\cos(x_2) + 0.7538\sin(x_2)\cos(x_3) + 0.2688\sin(x_3) - $ \\
 && $0.6707\cos(x_2) + 0.1723$ \\\hline
 7 & $4$ & $-0.6762x_0x_1 - 0.4155x_0\sin(x_3) + 0.3426x_1x_3 - 0.4999x_1 - 0.7566x_2x_3 + $ \\
 && $0.666x_2\sin(x_1) - 0.7283x_2 + 0.5425\sin(x_0)\sin(x_2) - 0.3538\cos(x_0) - 0.1851\cos(x_3) $ \\
 && $+ 0.8117$ \\\hline
 8 & $4$ & $0.5062x_0x_2 - 0.652x_0\sin(x_3) + 0.9153x_1x_3 - 0.7422x_1 + 0.0369x_2 - 0.2263x_3 $ \\
 && $-0.7665\sin(x_0) - 0.5118\sin(x_1)\cos(x_2) - 0.7336\sin(x_3)\cos(x_2)  $ \\
 && $- 0.1184\cos(x_0)\cos(x_1)+ 0.4495$ \\\hline
 9 & $4$ & $-0.7331x_0x_1 + 0.7149x_0x_3 - 0.937x_0\sin(x_2) - 0.8632x_1 + 0.5757x_3 + 0.7605\sin(x_0)$\\
 && $+ 0.3964\sin(x_1)\sin(x_3) + 0.3957\sin(x_2)\cos(x_1) + 0.5416\sin(x_2)\cos(x_3)  $ \\
 && $+ 0.7617\sin(x_2) + 0.8487$ \\
 10 & $4$ & $-0.1888x_0\sin(x_2) - 0.7688x_0 - 0.1821x_1x_3 + 0.7518x_1\cos(x_0) - 0.7683x_1\cos(x_2)$ \\
 && $- 0.3029x_1 + 0.5322x_2x_3 - 0.5291\sin(x_0)\cos(x_3) - 0.3467\sin(x_2) + 0.9045\sin(x_3)$ \\
 && $ - 0.8584$ \\
\hline
    \end{tabular}
\end{table}

\begin{table}[!t]
    \centering
    \caption{ Detailed expressions in Trigonometric datasets $(5,5,5)$ containing operators $\{\sin, \cos,+,-,\times\}$.}
    \label{tab:sincos-555}
    \begin{tabular}{c|cl}
    \toprule
        & \multicolumn{2}{c}{$\sin \cos (5, 5, 5)$} \\  
        Eq. ID & $n$ & Expression \\ \hline
 1 & $5$ & $-0.4156x_0x_1 - 0.1399x_2\cos(x_1) + 0.0438x_2 + 0.9508x_3\sin(x_1) + 0.2319x_3 $ \\
 && $- 0.6808x_4\cos(x_3) - 0.4468x_4 + 0.0585\sin(x_0) + 0.6224\cos(x_1) $ \\
 && $- 0.8638\cos(x_2)\cos(x_3)  + 0.959$ \\ \hline
 2 & $5$ & $0.5417x_0x_3 + 0.4562x_1\sin(x_3) + 0.0297x_1 + 0.3659x_2 + 0.638x_3\sin(x_2) - 0.0682x_3$ \\
 && $ - 0.7869\sin(x_0) - 0.7653\sin(x_3)\sin(x_4) - 0.5842\sin(x_4)\cos(x_0) + 0.1699\sin(x_4) $ \\
 && $+ 0.7374$ \\ \hline
 3 & $5$ & $-0.3012x_0x_3 - 0.2348x_0 - 0.8727x_1\cos(x_4) - 0.4123x_1 - 0.1288x_2\sin(x_4) + 0.2243x_3 $ \\
 && $+ 0.8088x_4 + 0.974\sin(x_2) + 0.3582\cos(x_2)\cos(x_3) - 0.5457\cos(x_3)\cos(x_4) - 0.2387$ \\ \hline
 4 & $5$ & $0.4818x_0\sin(x_4) - 0.8507x_0 - 0.0966x_1 - 0.9748x_2x_3 - 0.7097x_3x_4 - 0.4638x_3\cos(x_0) $ \\
 && $- 0.191x_4 - 0.0154\sin(x_2)\cos(x_4) + 0.16\cos(x_2) + 0.2525\cos(x_3) + 0.4967$ \\ \hline
 5 & $5$ & $-0.0203x_0\sin(x_4) + 0.6056x_0 + 0.8209x_1\cos(x_4) + 0.1261x_1 + 0.5846x_2\cos(x_1) $ \\
 && $- 0.92x_3\cos(x_4) - 0.7875x_4\cos(x_2) + 0.1995x_4 - 0.474\sin(x_2) + 0.5068\cos(x_3) $ \\
 && $- 0.8695$ \\ \hline
 6 & $5$ & $-0.8973x_0 - 0.8648x_1x_3 + 0.261x_1\sin(x_2) - 0.3262x_1 - 0.6032x_2\cos(x_0) - 0.6415x_2$ \\
 && $ - 0.0452x_3 - 0.8909x_4\cos(x_2) - 0.3137x_4 + 0.2508\sin(x_2)\sin(x_3) - 0.0501$ \\ \hline
 7 & $5$ & $0.7666x_0x_2 + 0.2223x_1x_2 + 0.3913x_1 - 0.8791x_2 - 0.3625x_3\sin(x_1) - 0.8499x_3\cos(x_0)$ \\
 && $ + 0.3701x_3 - 0.1628x_4\sin(x_0) - 0.3167x_4 - 0.8401\sin(x_0) - 0.6255$ \\ \hline
 8 & $5$ & $-0.6449x_0x_4 + 0.1764x_0 + 0.9283x_1\sin(x_0) + 0.1174x_1 - 0.7397x_2\cos(x_4) $ \\
 && $- 0.8867x_2 + 0.244x_3\cos(x_4) + 0.3295x_3 - 0.1753x_4 + 0.177\sin(x_2)\cos(x_3) - 0.381$ \\ \hline
 9 & $5$ & $0.105x_0\sin(x_2) + 0.8919x_3\sin(x_1) + 0.114x_3\cos(x_2) - 0.3825x_3 - 0.1461x_4\sin(x_0)$ \\
 && $ + 0.9091x_4\sin(x_1) - 0.6847\sin(x_0) + 0.9993\sin(x_2) - 0.1952\cos(x_1) + 0.6173\cos(x_4) $ \\
 && $- 0.4588$ \\ \hline
 10 & $5$ & $-0.7066x_0x_3 + 0.5513x_0 + 0.6577x_2x_3 - 0.7612x_4\sin(x_3) + 0.9294x_4 - $ \\
 && $0.2314\sin(x_2)\cos(x_4) + 0.5847\sin(x_2) + 0.5884\sin(x_3) + 0.3221\cos(x_1)\cos(x_3) $ \\
 && $+ 0.2867\cos(x_1) - 0.801$ \\ 
\hline
    \end{tabular}
\end{table}

\begin{table}[!t]
    \centering
    \caption{ Detailed expressions in Trigonometric datasets $(6,6,10)$ containing operators $\{\sin, \cos,+,-,\times\}$.}
    \label{tab:sincos-6610}
    \begin{tabular}{c|cl}
    \hline
        & \multicolumn{2}{c}{$\sin \cos (6, 6, 10)$} \\  
        Eq. ID & $n$ & Expression \\ \hline
 1 & $6$ & $0.7731x_0x_3 - 0.3506x_0x_5 + 0.5987x_1\cos(x_2) + 0.3325x_1 + 0.5694x_2x_5 + $ \\
 && $0.7267x_2\cos(x_0) - 0.0842x_3x_5 + 0.2041x_3\cos(x_1) + 0.4305x_4\sin(x_0) - 0.018x_4\sin(x_1) $ \\
 && $+ 0.8218x_4\cos(x_2) + 0.849x_4 + 0.8105x_5 - 0.532\sin(x_3) - 0.7069\cos(x_0) $ \\
 && $+ 0.2711\cos(x_2) + 0.7697$ \\ \hline
 2 & $6$ & $-0.4795x_0x_3 + 0.3408x_0x_5 - 0.2836x_0\sin(x_1) - 0.3653x_0\sin(x_4) - 0.4185x_2\sin(x_1)$ \\
 && $ - 0.5214x_2\cos(x_5) + 0.0565x_4 - 0.7513\sin(x_2)\sin(x_4) - 0.9501\sin(x_2)\cos(x_3) + $ \\
 && $0.2445\sin(x_2) + 0.6213\sin(x_3)\cos(x_1) - 0.1544\sin(x_3) - 0.7689\sin(x_5) + 0.1492\cos(x_0) $ \\
 && $+ 0.9059\cos(x_1) - 0.9596\cos(x_4)\cos(x_5) - 0.2943$ \\ \hline
 3 & $6$ & $-0.8018x_0x_3 - 0.6804x_0\cos(x_1) + 0.7706x_0\cos(x_2) + 0.4093x_0 - 0.0784x_1x_3 $ \\
 && $- 0.5774x_1x_4 - 0.321x_1\sin(x_2) + 0.7504x_1 + 0.9339x_2x_3 + 0.4978x_2x_5 + 0.1041x_4x_5 $ \\
 && $+ 0.1329x_4 - 0.4371x_5 + 0.776\sin(x_0)\sin(x_4) + 0.87\sin(x_2) - 0.0628\cos(x_3) + 0.8015$ \\ \hline
 4 & $6$ & $0.0879x_0\sin(x_2) + 0.2186x_0\cos(x_4) + 0.3456x_1 - 0.3001x_2x_5 + 0.1501x_2\cos(x_1)$ \\
 && $ - 0.4833x_3\sin(x_1) - 0.9198x_3\cos(x_4) + 0.3471x_4\sin(x_1) + 0.5014x_4\cos(x_2) $ \\
 && $+ 0.8932\sin(x_0)\cos(x_3) + 0.0212\sin(x_0)\cos(x_5) + 0.2926\sin(x_2) + 0.1891\sin(x_4) $ \\
 && $+ 0.3658\sin(x_5) - 0.3094\cos(x_0) + 0.1978\cos(x_3) - 0.3057$ \\ \hline
 5 & $6$ & $0.3458x_0x_1 - 0.3348x_0x_2 - 0.2813x_0x_5 - 0.9722x_0\sin(x_3) - 0.5237x_1\sin(x_4) $ \\
 && $- 0.827x_2\cos(x_5) - 0.8674x_3\sin(x_4) - 0.4151x_3\cos(x_5) - 0.473x_3 + 0.5282x_4\cos(x_0) $ \\
 && $- 0.9929x_4\cos(x_5) - 0.2496x_5 - 0.8814\sin(x_0) - 0.441\cos(x_1) - 0.5553\cos(x_2) $ \\
 && $+ 0.0649\cos(x_4) + 0.1151$ \\ \hline
 6 & $6$ & $0.7659x_0x_1 + 0.5859x_0x_4 + 0.5457x_0\sin(x_5) - 0.7902x_0 - 0.2567x_1\sin(x_4) $ \\
 && $+ 0.1881x_1\cos(x_3) - 0.9726x_2x_3 + 0.215x_2\sin(x_5) + 0.3626x_2 + 0.0368x_3\cos(x_4) $ \\
 && $+ 0.2092x_5\sin(x_3) + 0.4324x_5 + 0.8765\sin(x_0)\sin(x_3) - 0.7358\sin(x_1) + 0.6927\sin(x_3) $ \\
 && $+ 0.2547\sin(x_4) - 0.0298$ \\ \hline
 7 & $6$ & $0.4904x_0\sin(x_4) + 0.4611x_0\cos(x_5) - 0.9855x_1\sin(x_5) - 0.6639x_1$ \\
 && $ + 0.8481x_2\sin(x_0) + 0.7883x_3 + 0.3918x_4x_5 - 0.2737x_5\cos(x_3) + 0.56\sin(x_0)\sin(x_1) $ \\
 && $+ 0.455\sin(x_3)\cos(x_2) + 0.0465\sin(x_4) - 0.1309\cos(x_0) + 0.6162\cos(x_2)\cos(x_4) $ \\
 && $- 0.7968\cos(x_2) - 0.7076\cos(x_3)\cos(x_4) + 0.8198\cos(x_5) + 0.1901$ \\ \hline
 8 & $6$ & $-0.5442x_0x_3 - 0.6705x_0\sin(x_1) - 0.2059x_0\sin(x_5) - 0.1459x_0 + 0.2314x_1 $ \\
 && $- 0.5685x_2\sin(x_1) - 0.9793x_2 - 0.9266x_3\sin(x_5) - 0.5021x_3 + 0.6073x_4\sin(x_0) $ \\
 && $+ 0.4199x_4\sin(x_1) - 0.3315x_5\sin(x_2) + 0.4563x_5\cos(x_4) + 0.5843\sin(x_4) $ \\
 && $+ 0.3916\sin(x_5) + 0.1\cos(x_2)\cos(x_3) + 0.9413$ \\ \hline
 9 & $6$ & $0.9819x_0x_1 - 0.6092x_0x_3 + 0.4401x_0 + 0.2961x_1 - 0.2039x_2\cos(x_0) + 0.6163x_2 $ \\
 && $- 0.9343x_3\sin(x_1) - 0.5927x_4\sin(x_1) + 0.9x_5\cos(x_1) + 0.1278x_5\cos(x_2) - 0.4896x_5 $ \\
 && $+ 0.8773\sin(x_3)\cos(x_2) - 0.1269\sin(x_4)\cos(x_0) + 0.2907\sin(x_4) $ \\
 && $+ 0.5118\cos(x_2)\cos(x_4)  + 0.551\cos(x_3) + 0.1761$ \\ \hline
 10 & $6$ & $-0.9442x_0x_1 + 0.1517x_0\cos(x_4) + 0.1251x_0 - 0.3271x_1\sin(x_5) + 0.6235x_1\cos(x_3) $ \\
 && $- 0.6685x_1 + 0.4334x_2x_4 + 0.7275x_2\sin(x_0) - 0.8994x_2\sin(x_5) + 0.2948x_3x_4 $ \\
 && $- 0.2455x_3 + 0.0392x_5\sin(x_0) + 0.1698\sin(x_4) + 0.5975\cos(x_1)\cos(x_2) $ \\
 && $+ 0.1082\cos(x_2) + 0.0244\cos(x_5) + 0.0606$ \\ 
\hline
    \end{tabular}%
    
\end{table}

\begin{table}[!t]
    \centering
    \caption{ Detailed expressions in Trigonometric datasets $(8,8,12)$ containing operators $\{\sin, \cos,+,-,\times\}$.}
    \label{tab:sincos-8812}
    % \resizebox{\columnwidth}{!}{%
    \begin{tabular}{c|cl}
    \toprule
        & \multicolumn{2}{c}{$\sin \cos (8, 8, 12)$} \\  
        Eq. ID & $n$ & Expression \\ \hline
 1 & $8$ & $-0.3173x_0x_2 + 0.8872x_0x_4 + 0.6968x_0\cos(x_7) + 0.0237x_1\sin(x_0) + 0.5801x_3x_6 $ \\
 && $+ 0.2899x_3x_7 + 0.2512x_3\cos(x_5) - 0.3491x_3 + 0.2259x_4\sin(x_3) - 0.5449x_6 + 0.5309x_7$ \\
 && $ - 0.1076\sin(x_2) + 0.588\sin(x_4) + 0.6665\sin(x_5)\cos(x_0) - 0.8541\sin(x_7)\cos(x_6) $ \\
 && $+ 0.8479\cos(x_0) - 0.9064\cos(x_1)\cos(x_4) - 0.877\cos(x_1)\cos(x_7) - 0.8864\cos(x_1) $ \\
 && $+ 0.7559\cos(x_5) + 0.7899$ \\ \hline
 2 & $8$ & $0.2348x_0\sin(x_2) - 0.2717x_0 + 0.8024x_1x_2 + 0.9908x_1x_3 - 0.8918x_1x_5 - 0.9836x_4 $ \\
 && $- 0.2348x_5\sin(x_0) + 0.3617x_6x_7 + 0.2862x_6\sin(x_0) + 0.4972x_6\cos(x_5) + 0.9169x_6 $ \\
 && $- 0.2391x_7\sin(x_1) + 0.9162x_7\cos(x_5) + 0.3437x_7 + 0.052\sin(x_2)\cos(x_3) $ \\
 && $+ 0.4808\sin(x_3) + 0.3809\sin(x_4)\cos(x_3) - 0.3181\cos(x_1) - 0.946\cos(x_2) $ \\
 && $- 0.5821\cos(x_5) - 0.2309$ \\ \hline
 3 & $8$ & $0.1213x_0x_5 + 0.5367x_0x_6 - 0.255x_0\sin(x_2) + 0.8217x_0 - 0.6108x_1x_3 $ \\
 && $+ 0.4821x_2x_6 - 0.7619x_2\sin(x_4) - 0.8565x_2\cos(x_7) + 0.1538x_2 + 0.9386x_3\sin(x_4)$ \\
 && $ + 0.7035x_3\cos(x_6) + 0.48x_4\cos(x_6) - 0.7125x_5\cos(x_4) + 0.9182x_6\sin(x_5) $ \\
 && $- 0.7055x_6 + 0.7903\sin(x_1) + 0.3301\sin(x_4) - 0.8563\sin(x_5) - 0.8319\cos(x_3) $ \\
 && $+ 0.9053\cos(x_7) + 0.8513$ \\ \hline
 4 & $8$ & $-0.517x_0x_6 + 0.8709x_0\cos(x_3) + 0.445x_1\sin(x_2) - 0.7785x_1\cos(x_3) + 0.7564x_1 $ \\
 && $+ 0.1987x_2\cos(x_7) + 0.6675x_2 + 0.5181x_3x_4 + 0.3387x_4x_5 + 0.3192x_5 + 0.8468x_6\cos(x_3)$ \\
 && $ - 0.7186x_6\cos(x_5) - 0.8378x_6 - 0.2499x_7\cos(x_4) - 0.0502\sin(x_0) - 0.6406\sin(x_1)\cos(x_5) $ \\
 && $+ 0.6044\sin(x_2)\cos(x_6) - 0.067\cos(x_3) + 0.7396\cos(x_4) + 0.7652\cos(x_7) - 0.1012$ \\ \hline
 5 & $8$ & $-0.229x_0x_3 - 0.0595x_0x_6 + 0.7661x_0\cos(x_1) + 0.2791x_0\cos(x_4) + 0.0254x_0 $ \\
 && $- 0.7029x_1x_3 - 0.8571x_1x_5 - 0.0074x_1\sin(x_2) + 0.3041x_2 - 0.1725x_3x_5 - 0.158x_4\sin(x_1) $ \\
 && $- 0.6454x_4 - 0.8572x_5 - 0.8411x_6 - 0.5529x_7\sin(x_5) - 0.6246\sin(x_0)\cos(x_7) $ \\
 && $- 0.4132\sin(x_2)\sin(x_4) + 0.2489\sin(x_7) + 0.0314\cos(x_1) - 0.4135\cos(x_3) - 0.0282$ \\ \hline
 6 & $8$ & $-0.4509x_0x_6 - 0.2309x_1\cos(x_4) - 0.3714x_2\cos(x_0) + 0.1605x_3\cos(x_2) $ \\
 && $+ 0.5902x_4\sin(x_0) + 0.1579x_4\cos(x_3) + 0.3891x_4 + 0.0709x_5 - 0.0648x_6\cos(x_5) $ \\
 && $+ 0.9858x_6 + 0.2643x_7\cos(x_2) - 0.84\sin(x_0) + 0.8797\sin(x_1)\cos(x_5) + 0.8291\sin(x_1) $ \\
 && $+ 0.9734\sin(x_2)\sin(x_6) + 0.428\sin(x_2) - 0.7252\sin(x_3) + 0.0823\sin(x_4)\sin(x_6) $ \\
 && $- 0.3829\sin(x_7) - 0.2147\cos(x_1)\cos(x_7) + 0.0657$ \\ \hline
 7 & $8$ & $0.6837x_0 - 0.8837x_1x_2 - 0.5143x_1\sin(x_7) + 0.1197x_1\cos(x_0) - 0.4339x_1 $ \\
 && $+ 0.0717x_2x_7 - 0.5343x_3x_5 + 0.2918x_4\cos(x_5) - 0.35x_5 + 0.1734x_6\sin(x_2) - 0.6963x_7 $ \\
 && $+ 0.5374\sin(x_1)\sin(x_6) + 0.7663\sin(x_2) - 0.7015\sin(x_3)\cos(x_6) + 0.0821\sin(x_3) $ \\
 && $- 0.7863\sin(x_4) + 0.7635\cos(x_0)\cos(x_6) + 0.1298\cos(x_3)\cos(x_4) $ \\
 && $- 0.9813\cos(x_5)\cos(x_6) - 0.9848\cos(x_6) - 0.8544$ \\ \hline
 8 & $8$ & $0.9309x_0x_5 - 0.0939x_0\cos(x_6) - 0.2926x_0\cos(x_7) + 0.5092x_0 + 0.4184x_1\sin(x_2)$ \\
 && $ - 0.8222x_1\sin(x_7) + 0.256x_1\cos(x_6) - 0.5408x_1 + 0.4882x_2\sin(x_0) - 0.3633x_2 $ \\
 && $- 0.8636x_3x_4 + 0.2556x_3x_6 + 0.1037x_3 + 0.0417x_4x_7 + 0.9814x_4 - 0.2641x_6\cos(x_7) $ \\
 && $+ 0.4572x_6 - 0.8664x_7\cos(x_3) - 0.2983\cos(x_5) - 0.8532\cos(x_7) + 0.9698$ \\ \hline

    \end{tabular}%
    % }
\end{table}

\begin{table}[!t]
    \centering
    \caption{(Continued) Detailed expressions in Trigonometric datasets $(8,8,12)$ containing operators $\{\sin, \cos,+,-,\times\}$.}
    \label{tab:sincos-8812ct}
    % \resizebox{\columnwidth}{!}{%
    \begin{tabular}{c|cl}
    \toprule
        & \multicolumn{2}{c}{$\sin \cos (8, 8, 12)$} \\  
        Eq. ID & $n$ & Expression \\ \hline
 9 & $8$ & $0.292x_0\cos(x_2) - 0.1613x_0 + 0.3752x_1x_3 - 0.3503x_2\sin(x_1) + 0.1041x_2\sin(x_6) $ \\
 && $+ 0.6193x_4\cos(x_3) + 0.4697x_4 - 0.4962x_5x_7 - 0.006x_5\sin(x_3) + 0.2057x_5 $ \\
 && $- 0.7291x_6\sin(x_3) - 0.5308x_7\sin(x_6) - 0.1291x_7 + 0.871\sin(x_2)\sin(x_5) $ \\
 && $- 0.2832\sin(x_2) - 0.9686\sin(x_3) - 0.6923\sin(x_4)\cos(x_0) - 0.2232\cos(x_1) $ \\
 && $+ 0.1273\cos(x_4)\cos(x_6) + 0.2133\cos(x_6) - 0.8635$ \\ \hline
 10 & $8$ & $0.5465x_0\cos(x_3) - 0.6685x_0 - 0.8939x_1\sin(x_0) - 0.1214x_3x_7 - 0.4824x_3 $ \\
 && $- 0.9961x_4x_6 - 0.0657x_4\sin(x_2) - 0.6408x_4\cos(x_1) + 0.8089x_4 + 0.8806x_5\cos(x_4) $ \\
 && $- 0.5016x_5 - 0.0679x_6\sin(x_2) + 0.1362x_6 + 0.5724x_7\cos(x_1) + 0.3407x_7 $ \\
 && $- 0.713\sin(x_0)\sin(x_4) + 0.3745\sin(x_1) + 0.9728\sin(x_2)\cos(x_3) - 0.8915\sin(x_2) $ \\
 && $- 0.8401\sin(x_6)\cos(x_7) + 0.3912$ \\
 \hline

    \end{tabular}%
    % }
\end{table}

\begin{table}[!t]
    \centering
    \caption{Detailed expressions in Livermore2 datasets with input variables $n=4$.}
    \label{tab:livermore2-1}
    
    \begin{tabular}{c|cl}
    \toprule
        & \multicolumn{2}{c}{\textbf{Livermore2 ($n=4$)}} \\  
        Eq. ID & $n$ & Expression \\ \hline
Vars4-1 & $4$ & $x_0 - x_1x_2 - x_1 - 3x_3$ \\
 Vars4-2 & $4$ & $\sqrt{2}x_0\sqrt{x_1}x_3/x_2 + 1$ \\
 Vars4-3 & $4$ & $2x_0 + x_3 - 0.01 + x_2/x_1$ \\
 Vars4-4 & $4$ & $x_0 - x_3 - (-x_0 + \sin(x_0))^2/(x_0^2x_1^2x_2^2)$ \\
 Vars4-5 & $4$ & $x_0 + \sin(x_1/(x_0x_1^2x_3^2(-3.22x_1x_3^2 + 13.91x_1x_3 + x_2)/2 + x_1))^2$ \\
 Vars4-6 & $4$ & $-x_0 - 0.54\sqrt(x_3)\exp(x_0) + \exp(-2x_0)\cos(x_1)/x_2$ \\
 Vars4-7 & $4$ & $x_0 + x_2 + x_3 + \cos(x_1)/\log(x_1^2 + 1)$ \\
 Vars4-8 & $4$ & $x_0(x_0 + x_3 + \sin((-x_0\exp(x_2) + x_1)/(-4.47x_0^2x_2 + 8.31x_2^3 + 5.27x_2^2))) - x_0$ \\
 Vars4-9 & $4$ & $x_0 - x_3 + \cos(x_0(x_0 + x_1)(x_0^2x_1 + x_2) + x_2)$ \\
 Vars4-10 & $4$ & $x_0(x_3 + (\sqrt{x_1} - \sin(x_2))/x_2) + x_0$ \\
 Vars4-11 & $4$ & $2x_0 + x_1(x_0 + \sin(x_1x_2)) + \sin(2/x_3)$ \\
 Vars4-12 & $4$ & $x_0x_1 + 16.97x_2 - x_3$ \\
 Vars4-13 & $4$ & $x_3(-x_2 - \sin(x_0^2 - x_0 + x_1))$ \\
 Vars4-14 & $4$ & $x_0 + \cos(x_1^2(-x_1 + x_2 + 3.23) + x_3)$ \\
 Vars4-15 & $4$ & $x_0(x_1 + \log(x_2 + x_3 + \exp(x_1^2) - 0.28/x_0)) - x_2 - x_3/(2x_0x_2)$ \\
 Vars4-16 & $4$ & $-x_0^2\sqrt{x_1} + x_2(-x_3 + 1.81/x_2) + \exp(x_1) - 2.34x_3/x_0$ \\
 Vars4-17 & $4$ & $x_0^2 - x_1 - x_2^2 - x_3$ \\
 Vars4-18 & $4$ & $x_0 - x_3\exp(x_0) + 2.96\sqrt{0.36x_1^2 + x_1x_2^2 + 0.94} + \log(-x_0 + x_1 + 1) + \sin(2x_1 + x_2)$ \\
 Vars4-19 & $4$ & $(x_0^3x_1 - 2.86x_0 + x_3)/x_2$ \\
 Vars4-20 & $4$ & $x_0 + x_1 + 6.21 + 1/(x_2x_3 + x_2 + 2.08)$ \\
 Vars4-21 & $4$ & $x_0(x_1 - x_2 + x_3) + 2x_3$ \\
 Vars4-22 & $4$ & $2x_0 - x_1x_2 + x_1\exp(x_0) - x_3$ \\
 Vars4-23 & $4$ & $-x_0/x_1 - 2.23x_1x_2 + x_1 - 2.23x_2/\sqrt{x_3} - 2.23\sqrt{x_3} + \log(x_0)$ \\
 Vars4-24 & $4$ & $-4.81x_0x_1\log(x_0) + x_0 + \sqrt{x_3} + \log(x_2)$ \\
 Vars4-25 & $4$ & $0.38 + (-x_0/x_3 + \cos(2x_0x_2/(x_3(x_0 + x_1x_2)))/x_3)/x_1$ \\
\hline
    \end{tabular}
\end{table}

\begin{table*}[!t]
    \centering
    \caption{ Detailed expressions in Livermore2 datasets  with input variables $n=5$.}
    \label{tab:livermore2-2}
    \begin{tabular}{c|cl}
    \toprule
        & \multicolumn{2}{c}{\textbf{Livermore2 ($n=5$)}} \\  
        Eq. ID & $n$ & Expression \\ \hline
Vars5-1 & $5$ & $-x_0 + x_1 - x_2 + x_3 - x_4 - 4.75$ \\
 Vars5-2 & $5$ & $x_2(x_0 + x_4 + 0.27/(x_2^2 + (x_1 + x_3)/(x_0x_1 + x_1)))$ \\
 Vars5-3 & $5$ & $2x_0x_1x_2 + x_4 - \sin(x_0\log(x_1 + 1) - x_0 + x_3)$ \\
 Vars5-4 & $5$ & $x_1 + x_2x_3 + x_4^2 + \sin(x_0)$ \\
 Vars5-5 & $5$ & $\sqrt{x_2} + x_4 + \log(x_1 + x_3) + 0.36\log(x_0x_1 + 1.0)$ \\
 Vars5-6 & $5$ & $x_0x_3 + x_0 + x_1 + x_4 + \sqrt{0.08x_0/(x_2x_4) + x_2}$ \\
 Vars5-7 & $5$ & $x_0x_4 - x_0/(x_1 + x_2 + x_3 + 8.05) + \sqrt(x_0x_1)\cos(x_0)$ \\
 Vars5-8 & $5$ & $\sqrt{x_1}x_2 - x_3 - (0.07x_0 + 0.07(x_0 - x_1)\sqrt{x_1 + 0.99})\cos(x_4)$ \\
 Vars5-9 & $5$ & $x_0(x_2 + (x_0 + x_1)/(x_1x_3 + x_4))$ \\
 Vars5-10 & $5$ & $x_0(-0.25x_0x_2x_3 + x_1 - 8.43x_3x_4)\sin(x_2 + 1)/x_3 + x_3x_4$ \\
 Vars5-11 & $5$ & $-x_1 - x_3^2 + 0.47\sqrt{x_0x_2} + \sqrt{x_0x_2 + x_4} + x_4/x_2 - 1/\sqrt(x_1)$ \\
 Vars5-12 & $5$ & $x_0(x_1 - 1/(x_2(x_3 + x_4)))$ \\
 Vars5-13 & $5$ & $\sqrt{x_0x_4}(x_1 - 1.52) - \cos(4.03x_2 + x_3)$ \\
 Vars5-14 & $5$ & $-x_0/(x_1x_4) + \cos(x_0x_2x_3\exp(-x_1))$ \\
 Vars5-15 & $5$ & $x_2 - x_3 + x_4 + \sqrt{x_1x_4} + \log(x_0 + 1)/\log(11.06x_1x_4 + 1) - \cos(x_1)$ \\
 Vars5-16 & $5$ & $x_1 + 0.33x_4(x_0/(x_0^2 + x_1) + x_2x_3)$ \\
 Vars5-17 & $5$ & $x_0 - \sin(x_1) + \sin(x_2) - \cos(-x_1 + \sqrt{x_3} + x_4) + 0.78$ \\
 Vars5-18 & $5$ & $x_0x_1 - x_3 - (x_2\sqrt{1/(x_0(x_2 + x_3))} - 1.13/x_2)/x_4$ \\
 Vars5-19 & $5$ & $4.53x_0x_1 + x_0 - x_0\cos(\sqrt{x_1})/x_1 - x_2 - x_3 - x_4$ \\
 Vars5-20 & $5$ & $x_1\sin(x_0 - 4.81)/(0.21x_4 - 0.21\exp(x_4) - 0.21\log(x_2 + x_3 + 1)) - \exp(x_0 + x_4)$ \\
 Vars5-21 & $5$ & $2x_0\sqrt{x_3} + x_2 + \exp(x_0x_1)\cos(x_0x_2x_3) - \log(x_2 + 3.49)/x_4$ \\
 Vars5-22 & $5$ & $-x_1(\sin(x_2) - \log(x_0x_4/(x_1^2 + x_3) + 1)/x_3) - x_1 + x_2 - 0.73$ \\
 Vars5-23 & $5$ & $x_0(x_1/(x_2 + \sqrt{x_1(x_3 + x_4)}(-x_2 + x_3 + 1)) - x_4)$ \\
 Vars5-24 & $5$ & $\sqrt{x_0} - x_1x_4 + x_1(-x_0 + x_3\cos(\sqrt{x_2} + x_2) - (x_1 + 7.84x_2^2x_4)/x_4) + x_1/x_2$ \\
 Vars5-25 & $5$ & $x_0 + \log(x_0(-3.57x_0^2x_1 + x_0 + x_1 + x_2\log(-x_0x_3\sin(x_2)/x_4 + x_2)))$ \\
         \bottomrule
    \end{tabular}
\end{table*}

\begin{table*}[!t]
    \centering
    \caption{ Detailed expressions in Livermore2 dataset  with input variables $n=6$.}
    \label{tab:livermore2-3}
    \begin{tabular}{c|cl}
\toprule
        & \multicolumn{2}{c}{\textbf{Livermore2 ($n=6$)}} \\  
        Eq. ID & $n$ & Expression \\ \hline
Vars6-1 & $6$ & $x_0 - x_2 - x_5 + \sqrt(x_0^2 + x_1)(x_0 + x_3 + x_4)$ \\
 Vars6-2 & $6$ & $x_0(2x_1 + x_1/x_2 + x_3 + \log(x_0x_4x_5 + 1.0))$ \\
 Vars6-3 & $6$ & $\sqrt(x_1 + x_4) - \sqrt(x_5 + x_2^4x_3^4/(x_0x_1^4))$ \\
 Vars6-4 & $6$ & $x_0(x_1(x_0^2 + x_0) - x_1 + x_2^2 - x_2 - x_4 - x_5 - \sin(x_3) - \cos(x_3))^2$ \\
 Vars6-5 & $6$ & $x_1\sqrt(x_0x_1)(x_0x_2 - x_2 - x_3) + x_4 + x_5$ \\
 Vars6-6 & $6$ & $(x_0/(x_1x_2 + 2\log(\cos(x_0) + 2)) - x_1x_3 + \sin((x_1x_3 + x_4)/x_5) + \cos(x_2))\log(x_0)$ \\
 Vars6-7 & $6$ & $x_0^(3/2) - x_5^2 + \sin((x_0\exp(-x_1) - x_3(x_1 + x_2^2))/(x_1 + x_4))$ \\
 Vars6-8 & $6$ & $x_0 + x_1^2 + 0.34x_2x_4 - x_3 + x_5$ \\
 Vars6-9 & $6$ & $x_0x_3 + x_1 - 4x_4^2\log(x_1)/(x_0x_2 - x_1^2) - x_5 + \exp(13.28x_2x_5) - \log(x_2 + 0.5)$ \\
 Vars6-10 & $6$ & $x_0 + 61.36x_1^2 + x_1/(x_0x_2(x_3 - \cos(x_3(2x_0x_1x_5/x_4 + x_4))))$ \\
 Vars6-11 & $6$ & $(x_0 + x_0/(x_1 + x_3(8.13x_0^2x_5 + x_0x_1x_2 + 2x_1 + x_4 + x_5)))^2$ \\
 Vars6-12 & $6$ & $(1.41\sqrt(x_0) - x_1 - x_2/\sqrt(x_3(8.29x_0x_2^2 + x_0x_4) + x_3 + x_5))/x_5$ \\
 Vars6-13 & $6$ & $x_0 + x_4 + 0.21\sqrt(x_0/(x_1^2x_2^2\sqrt(x_5)(\sqrt(x_2) + x_2 + 2x_5 + (x_1 + x_3 + x_4)/x_4)))$ \\
 Vars6-14 & $6$ & $-2.07x_5 + \log(x_1 - x_5 - \sqrt(x_2(x_4 + \log(-x_0 + x_4 + 1))/x_3))$ \\
 Vars6-15 & $6$ & $x_0(x_0 + \cos(x_1^2x_2x_3(x_4 - 0.43x_5^2)))/x_3$ \\
 Vars6-16 & $6$ & $-\sqrt(x_0) - x_0 + x_1 - x_3 - x_4 - \sqrt(x_5/x_2) - 3.26$ \\
 Vars6-17 & $6$ & $x_0/(x_1x_3) + (-x_4 + 2\log(x_5)\cos(2x_1 + x_2^2 - x_3))(129.28x_0^2x_1^2 + x_2)$ \\
 Vars6-18 & $6$ & $\sqrt(x_4)(2x_0 + \cos(x_0(x_2x_3\exp(x_0x_1) + x_2 - \log(x_2 + 0.5) - 3.49))/x_5)$ \\
 Vars6-19 & $6$ & $x_0 + 2x_1 + x_2 + x_3 - x_4 + 0.84\sqrt(x_2x_5) + \log(x_2 + 0.5) + \exp(x_1)/(x_1 - x_3)$ \\
 Vars6-20 & $6$ & $x_0 - 0.97x_0/(x_4 - x_5(x_0x_3 + x_5)) - x_1 + x_2 + \sin(x_0^2)/x_0$ \\
 Vars6-21 & $6$ & $x_0 + x_2 + x_3 - \log(x_5 + 0.5)\sin(3.47x_1)/x_4 + 25.56\exp(x_4)\sin(x_1)/x_1$ \\
 Vars6-22 & $6$ & $x_0 + (x_3 - \sin(0.22x_2 - 0.22x_3 + 0.22)\cos(x_5))\cos(x_1 + 2.27x_4)$ \\
 Vars6-23 & $6$ & $x_0(-x_5 + 1.88\sqrt(0.71x_0 + x_1)) + x_0 + 0.28x_2 + x_3 - 0.28x_3/x_4 + 2\log(x_0)$ \\
 Vars6-24 & $6$ & $0.24x_1 + 1.42\sqrt(x_2)/(x_5\sqrt(x_3 + x_4)) + \sin(x_0)/x_5$ \\
 Vars6-25 & $6$ & $x_0 - x_1^2 - x_2 + x_4\cos(x_2) + x_4 + x_5 - 2.19\sqrt(x_2 + 0.44/x_3)$ \\

\hline
    \end{tabular}
\end{table*}

\begin{table*}[!ht]
    \centering
    \caption{Detailed equations in Feynman datasets ($n=4$). $n$ stands for the number of maximum variables.}
    \label{tab:Feynman-4}
    \begin{tabular}{l|cl}
\toprule
        & \multicolumn{2}{c}{\textbf{Feynman} ($n=4$)} \\  
        Eq. ID & $n$ & Expression \\ \hline
I.8.14 & $4$ & $\sqrt{(x_0 - x_1)^2 + (x_2 - x_3)^2}$ \\
I.13.4 & $4$ & $0.5x_0(x_1^2 + x_2^2 + x_3^2)$ \\
I.13.12 & $4$ & $6.6743e-11x_0x_1(-1/x_3 + 1/x_2)$ \\
I.18.4 & $4$ & $(x_0x_1 + x_2x_3)/(x_0 + x_2)$ \\
I.18.16 & $4$ & $x_0x_1x_2\sin(x_3)$ \\
I.24.6 & $4$ & $0.25x_0x_3^2(x_1^2 + x_2^2)$ \\
I.29.16 & $4$ & $\sqrt{x_0^2 + 2x_0x_1\cos(x_2 - x_3) + x_1^2}$ \\
I.32.17 & $4$ & $0.0035\pi x_0^2x_1^2x_2^4/(x_2^2 - x_3^2)^2$ \\
I.34.8 & $4$ & $x_0x_1x_2/x_3$ \\
I.40.1 & $4$ & $x_0\exp(-7.10292768111229e+23x_1x_2/x_3)$ \\
I.43.16 & $4$ & $x_0x_1x_2/x_3$ \\
I.44.4 & $4$ & $1.38e-23x_0x_1\log(x_2/x_3)$ \\
I.50.26 & $4$ & $x_0(x_3\cos(x_1x_2)^2 + \cos(x_1x_2))$ \\
II.11.20 & $4$ & $2.41e+22x_0x_1^2x_2/x_3$ \\
II.34.11 & $4$ & $x_0x_1x_2/(2x_3)$ \\
II.35.18 & $4$ & $x_0/(\exp(7.24e+22x_1x_2/x_3) + \exp(-7.24e+22x_1x_2/x_3))$ \\
II.35.21 & $4$ & $x_0x_1\tanh(7.24e+22x_1x_2/x_3)$ \\
II.38.3 & $4$ & $x_0x_1x_2/x_3$ \\
III.10.19 & $4$ & $x_0\sqrt{x_1^2 + x_2^2 + x_3^2}$ \\
III.14.14 & $4$ & $x_0(\exp(7.24e+22x_1x_2/x_3) - 1)$ \\
III.21.20 & $4$ & $-x_0x_1x_2/x_3$ \\
BONUS.1 & $4$ & $3.32e-57x_0^2x_1^2/(x_2^2\sin(x_3/2)^4)$ \\
BONUS.3 & $4$ & $x_0(1 - x_1^2)/(x_1\cos(x_2 - x_3) + 1)$ \\
BONUS.11 & $4$ & $4x_0\sin(x_1/2)^2\sin(x_2x_3/2)^2/(x_1^2\sin(x_3/2)^2)$ \\
BONUS.19 & $4$ & $-1872855580.36049(8.07e+33x_0/x_1^2 + 8.98e+16x_2^2(1 - 2x_3))/\pi$ \\
\hline
\end{tabular}
\end{table*}

\begin{table*}[!ht]
    \centering
    \caption{Detailed equations in Feynman datasets ($n=5$). $n$ stands for the number of maximum variables.}
    \label{tab:Feynman-5}
    \begin{tabular}{l|cl}
\toprule
& \multicolumn{2}{c}{\textbf{Feynman} ($n=5$)} \\  
        Eq. ID & $n$ & Expression \\ \hline
    I.12.11 & $5$ & $x_0(x_1 + x_2x_3\sin(x_4))$\\
    II.2.42& $5$ & $x_0x_3(x_1 - x_2)/x_4$\\
    II.6.15a& $5$ & $84707476846.623x_0x_1\sqrt{(x_3^2 + x_4^2)}/(\pi x_2^5)$\\
    II.11.3& $5$ & $x_0x_1/(x_2(x_3^2 - x_4^2))$\\
    II.11.17& $5$ & $x_0(7.24e+22x_1x_2\cos(x_3)/x_4 + 1)$\\
    II.36.38& $5$ & $7.24e+22x_0x_1/x_2 + 9.10e+16x_0x_3x_4/x_2$\\
    III.9.52& $5$ & $1.21e+34\pi x_0x_1\sin(x_2(x_3 - x_4)/2)^2/(x_2(x_3 - x_4)^2)$\\
    bonus.4& $5$ & $\sqrt{2}\sqrt{(x_1 - x_2 - x_3^2/(2x_0x_4^2))/x_0}$\\
bonus.12& $5$ & $x_0(-x_0x_2^3x_4/(x_2^2 - x_4^2)^2 + 4pix_1x_3x_4)/(4\pi x_1x_2^2)$\\
bonus.13& $5$ & $x_1/(4\pi x_0\sqrt{x_2^2 - 2x_2x_3\cos(x_4) + x_3^2})$\\ 
bonus.14& $5$ & $x_0(-x_2 + x_3^3(x_4 - 1)/(x_2^2(x_4 + 2)))\cos(x_1)$\\
bonus.16& $5$ & $x_1x_4 + 8.98e+16\sqrt{x_3^2 + 1.11e-17(x_0 - x_1x_2)^2}$\\
\hline
\end{tabular}
\end{table*}

\clearpage
\subsection{Baseline Implementation}\label{apx:baseline}
\paragraph{\method} Our method is implemented on top of the \textbf{GP} following Algorithm~\ref{alg:gp}. See the codebase for more details.

\paragraph{GP, CVGP} The implementation of the two algorithms are available at\footnote{\url{https://github.com/jiangnanhugo/cvgp}}.

\begin{figure}[!t]
    \centering
    \includegraphics[width=1.0\linewidth]{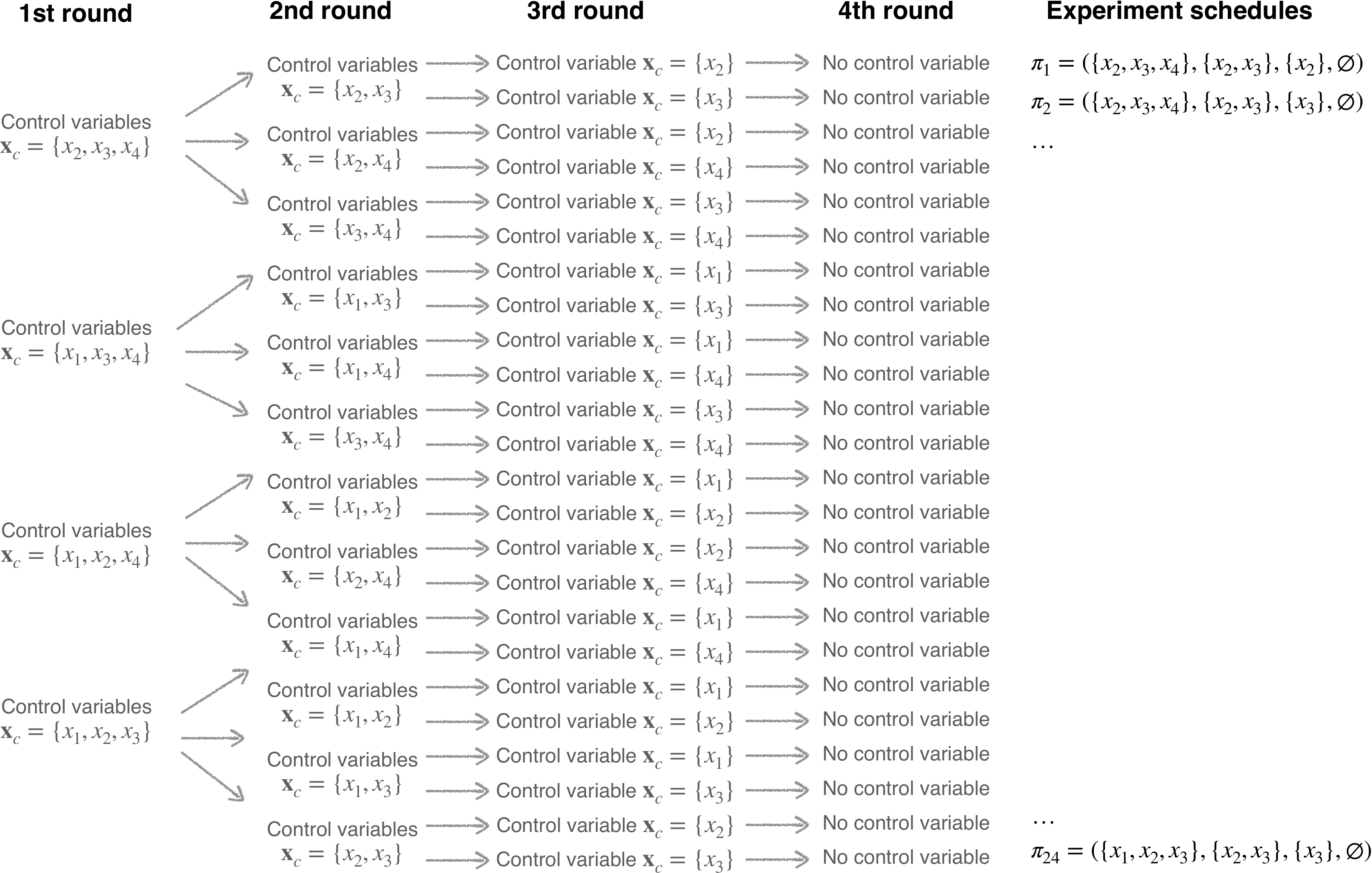}
    \caption{Execution of CVGP with all experiment schedules for expressions with $4$ input variables.}
    \label{fig:all-cvgp}
\end{figure}

\paragraph{CVGP (all experiment schedules).} We implement this baseline using a tree data structure to maintain all the experiment schedules. In Figure~\ref{fig:all-cvgp}, we visualize the whole tree over all $24$ possible experiment schedules $\pi_1,\ldots,\pi_{24}$. This will be more time efficient than just launching the CVGP algorithm for $4!=24$ times because we do not need to rerun 1st round and 2nd rounds. Note that at every round for a fixed set of control variables $\mathbf{x}_c$, it corresponds to running the GP algorithm for $\mathtt{\#Gen}$ generations over  $\#M$ expressions in the population pools.

\paragraph{Eureqa} This algorithm is currently maintained by the DataRobot webiste\footnote{\url{https://docs.datarobot.com/en/docs/modeling/analyze-models/describe/eureqa.html}}. We use the python API provided \footnote{\url{https://pypi.org/project/datarobot/}} to send the training dataset to the DataRobot website and collect the predicted expression after 30 minutes. This website only allows us to execute their program under a limited budget. Due to budgetary constraints, we were only able to test the datasets for the noiseless settings. 
For the Eureqa method, the fitness measure function is negative RMSE. We generated large datasets of size $10^5$ in training each benchmark.

\paragraph{DSR, PQT, GPMeld} These algorithms are evaluated based on an implementation in\footnote{\url{https://github.com/brendenpetersen/deep-symbolic-optimization}}. For every ground-truth expression, we generate a dataset of sizes $10^5$ training samples. Then we execute all these baselines on the dataset with the configurations listed in Table~\ref{tab:apx-configuration}.  
% For the four baselines (\textit{i.e.}, PQT, VPG, DSR, GPMeld), the reward function is INV-NRMSE, which is defined as $\frac{1}{1+\text{NRMSE}}$. 

 The official implementation of Symbolic Physics Learner (SPL)\footnote{\url{https://github.com/isds-neu/SymbolicPhysicsLearner}} does not support discovering equations with constants. Thus SPL is not considered in this research.

\subsection{Hyper-parameter Configuration}\label{apx:hyper-parameter-config}
We list the major hyper-parameter setting for all the algorithms in Table~\ref{tab:apx-configuration}. Note that if we use the default parameter settings, the GPMeld algorithm takes more than 1 day to train on one dataset. 
Because of such slow performance, we cut the number of genetic programming generations in GPMeld by half to ensure fair comparisons with other approaches.

\begin{table}[!ht]
    \centering
     \caption{Major hyper-parameters settings for all the algorithms considered in the experiment.}
    \label{tab:apx-configuration}
    \resizebox{\columnwidth}{!}{%
    \begin{tabular}{r|ccccccc} \toprule
          & \method &CVGP & GP&  DSR & PQT & GPMeld &Eureqa\\\hline
         Reward function & NegMSE & NegMSE &NegMSE & InvNRMSE & InvNRMSE & InvNRMSE & NegRMSE\\ 
        Training set size & $25,600$ & $25,600$ & $25,600$ & $$ 50, 000$$ & $ 50, 000$ & $ 50, 000$ & $ 50, 000$ \\
         Testing set size & $256$ & $256$ & $256$ & $256$ & $256$ & $ 256$ & $ 256$ \\
         Batch size & $256$ & $256$ & $256$ & $1024$ & $1024$ & $1024$ & $N/A$\\
      \#CPUs for training& 1 & 1  & 1 & 4 & 4 & 4 & 1\\
  $\epsilon$-risk-seeking policy  & N/A & N/A  &  N/A & 0.02 & N/A & 0.02 & N/A\\ \hline
        \#genetic generations & 100 & 100 & 100 & N/A & N/A  & 20  &10,000\\
        \#Hall of fame &  25 & 25 & 25  & N/A & N/A & 25 & N/A \\
        Mutation Probability & {0.5} & {0.5} & 0.5 & N/A & N/A &0.5 &N/A \\
        Mating Probability & {0.5} & {0.5} & 0.5 & N/A & N/A &0.5 &N/A  \\
        \hline
        Max training time& \multicolumn{7}{c}{48 hours}\\
        \bottomrule
    \end{tabular}%
    }
\end{table}

\end{document}